\documentclass[journal,twoside,web]{ieeecolor}
\usepackage{generic}
\usepackage{cite}
\usepackage{amsmath,amssymb,amsfonts}
\usepackage{algorithm}
\usepackage{algpseudocode}
\usepackage{graphicx}
\usepackage{hyperref}
\usepackage{textcomp}
\usepackage{makecell} 
\usepackage{xcolor}
\colorlet{subsectioncolor}{black}
\usepackage{adjustbox}
\usepackage{booktabs}
\usepackage{caption}
\usepackage{multirow}

\usepackage{multirow}

\makeatletter
\def\ps@titlepagestyle{%
  \def\@oddhead{}%
  \def\@evenhead{}%
  \def\@oddfoot{}%
  \def\@evenfoot{}%
}
\def\ps@headings{%
  \def\@oddhead{}%
  \def\@evenhead{}%
  \def\@oddfoot{}%
  \def\@evenfoot{}%
}
\makeatother
\begin{document}

\title{Hierarchical Spatio-Channel Clustering for Efficient Model Compression in Medical Image Analysis}

\author{Sisipho Hamlomo, Marcellin Atemkeng, Habte Tadesse Likassa, Blaise Ravelo, Senior Member, IEEE, Thierry Bouwmans, Sébastien Lalléchère, Senior Member, IEEE, Antoine Vacavant, and Ding-Geng Chen
\thanks{Manuscript received April 23, 2026; revised XX XX, 2026; accepted XX XX, 2026. This work was supported in part by the Department of Higher Education and Training through the University Staff Doctoral Programme and in part by the National Research Foundation of South Africa under Grant CSRP23040990793.}%
\thanks{S. Hamlomo is with the Department of Mathematics, Rhodes University, Makhanda 6140, South Africa, and also with the Department of Statistics, Rhodes University, Makhanda 6140, South Africa (corresponding author, e-mail: s.hamlomo@ru.ac.za). M. Atemkeng is with the Department of Mathematics, Rhodes University, Makhanda 6140, South Africa, and also with the National Institute for Theoretical and Computational Sciences, Stellenbosch 7600, South Africa (corresponding author, e-mail: m.atemkeng@ru.ac.za).}%
\thanks{H. T. Likassa is with the College of Health Solutions, Arizona State University, Phoenix, AZ 85004 USA, and also with the Department of Statistics, College of Natural and Computational Sciences, Addis Ababa University, Addis Ababa, Ethiopia. D.-G. Chen is with the College of Health Solutions, Arizona State University, Phoenix, AZ 85004 USA, and also with the Department of Statistics, University of Pretoria, Pretoria, South Africa.}%
\thanks{A. Vacavant is with Université Clermont Auvergne, Clermont Auvergne INP, CNRS, Institut Pascal, Clermont--Ferrand, France. T. Bouwmans is with Laboratoire MIA, La Rochelle Université, La Rochelle, France. S. Lalléchère is with the Association Française de Science des Systèmes (AFSCET), Fontenay-aux-Roses, France.}%
\thanks{B. Ravelo is with the Computer Vision, Institute of Cognitive Science, Osnabrück University, Osnabrück D-49090, Germany, and also with the Nanjing University of Information Science and Technology, Nanjing 210044, China.}%
}

\maketitle

\begin{abstract}
Convolutional neural networks (CNNs) have become increasingly difficult to deploy in resource-constrained environments due to their large memory and computational requirements. Although low-rank compression methods can reduce this burden, most existing approaches compress spatial and channel redundancy independently and therefore do not fully exploit the localised structure present within convolutional feature maps. This paper proposes a hierarchical spatio-channel low-rank compression mathematical framework for CNNs that adaptively exploits redundancy across spatial regions and channel activations. Unlike conventional low-rank methods, which apply a uniform decomposition across an entire layer, the proposed approach first partitions feature maps into spatial regions, then groups channels according to their co-activation patterns within each region, and finally applies rank-adaptive singular value decomposition (SVD) to each resulting spatio-channel cluster. The method is evaluated on an AlexNet-based brain tumour MRI classification model and compared with Global SVD and Tucker decomposition under \(3\times\) and \(6\times\) compression budgets. Our proposed method outperforms both Global SVD and Tucker decomposition, reducing FLOPs from \(8.21\,\mathrm{G}\) to \(1.55\,\mathrm{G}\), i.e., \(81.1\%\) reduction, achieving a \(1.38\times\) inference speed-up, and increasing classification accuracy from \(87.76\%\) to \(89.80\%\). The method also improves the macro \(F_1\)-score and yields better performance on challenging classes such as meningioma. A hyper-parameter trade-off analysis further demonstrates that the proposed framework provides a range of Pareto-optimal configurations, enabling flexible control over the balance between compression and predictive performance. Moderate spatial and channel clustering settings, combined with adaptive rank selection, consistently yield strong results, indicating that the framework is robust across a broad range of parameter settings. Bootstrap standard errors are reported for all classification metrics to quantify the uncertainty associated with the finite test set.
\end{abstract}

\begin{IEEEkeywords}
Model Compression, Low-Rank Approximation, Spatio-Channel Clustering, Medical Imaging
%Enter key words or phrases in alphabetical order, separated by commas. Using the IEEE Thesaurus can help you find the best standardized keywords to fit your article. Use the thesaurus access request form for free access to the IEEE Thesaurus: \underline{https://www.ieee.org/publications/services/thesaurus-acce}\\
%\underline{ss-page.com.}
\end{IEEEkeywords}

\section{Introduction}
Deep neural networks have increased in size over time, leading to higher memory usage and computational costs \cite{hoefler2021sparsity,bartoldson2023compute,menghani2023efficient}. Modern convolutional and transformer-based architectures often contain hundreds of millions of parameters, making them challenging to deploy in resource-constrained environments \cite{bergstra2013making, cheng2018model, shoeybi2019megatron, dalvi2020analyzing}. A significant factor behind this inefficiency is redundancy in the learned representations: filters, channels, and feature maps frequently encode overlapping information that can be represented more compactly while minimising accuracy loss \cite{cheng2018model, dalvi2020analyzing, chen2023survey}. Developing compression techniques that effectively remove this redundancy while retaining predictive performance has therefore become a key area of research.

One approach to exploiting this redundancy is low-rank matrix approximation (LORMA). By factorising weight matrices or tensors into low-rank components, LORMA reduces both storage and computation while preserving the dominant structure. Early studies in computer vision demonstrated that convolutional filters could be approximated by low-rank spatial bases or tensor decompositions, yielding significant speedups with minimal loss in accuracy \cite{jaderberg2014speeding, ioannou2015training}. In natural language processing, similar ideas have been applied to compress embedding layers, with methods such as online low-rank factorisation \cite{acharya2019online}, block-wise approximation guided by token frequencies \cite{chen2018groupreduce}, projective clustering into multiple subspaces \cite{maalouf2020deep}, and robust alternatives to classical SVD \cite{tukan2021no}. More recent work has emphasised adaptivity, with strategies that learn layer-specific ranks \cite{idelbayev2020low}, allocate ranks dynamically \cite{gao2024adaptive}, or perform automated decomposition search \cite{liebenwein2021compressing}. Hardware-aware designs such as MCUBERT \cite{yang2024mcubert} further highlight the potential of low‑rank methods in resource‑constrained edge devices. These contributions illustrate the broad applicability of low-rank approximation across domains and tasks. While these approaches demonstrate the effectiveness of low-rank compression, many existing methods compress spatial or channel redundancy independently, or apply uniform tensor decompositions that do not explicitly adapt to localised variation. Deep feature maps can exhibit complex spatio-channel correlations, in which localised activation patterns are associated with the co-activation of specific channel subsets \cite{lin2017espace,li2023scconv}. Ignoring this joint structure may limit compression efficiency and risk discarding meaningful information. Although methods such as subspace exploration in feature maps \cite{wang2018exploring} and adaptive decomposition frameworks \cite{liebenwein2021compressing} address related aspects, they do not explicitly model hierarchical dependencies across both spatial regions and channel groups prior to decomposition.

The idea of combining clustering with low‑rank approximation has recently been explored for medical image compression, where similar image patches or multimodal regions are grouped before applying low‑rank matrix approximation within each cluster \cite{hamlomo2026clustering}. This motivates the present work. However, rather than working directly on image patches, we transfer the idea to CNNs: we cluster spatial regions of feature maps and group channels by their co‑activation patterns before low‑rank decomposition.

Building on this idea, this paper introduces a hierarchical spatio-channel low-rank compression framework that captures this joint structure. The method first clusters feature maps into spatial regions to identify localised activation patterns, then groups channels within each region according to their co-activation patterns, and finally applies truncated SVD with adaptive rank selection to each spatio-channel cluster. By aligning compression with both spatial and channel coherence, the approach produces compact representations while maintaining predictive performance.

Efficient and reliable compression is particularly relevant in medical image analysis, where models are often deployed in environments with limited computational resources and where preserving diagnostically relevant structures is essential. Evaluation is performed on a brain tumour MRI classification task.

The main contributions of this work are as follows:
\begin{itemize}
    \item We propose a hierarchical spatio-channel low-rank compression framework that jointly models spatial and channel-wise redundancy in convolutional feature maps through activation-guided clustering and adaptive rank allocation. By aligning low-rank approximation with localised activation patterns, the proposed method preserves discriminative features while effectively reducing redundancy across layers.
    
    \item We provide a detailed empirical analysis, including layer-wise compression behaviour and hyper-parameter trade-offs, showing that the proposed framework enables flexible and robust efficiency--accuracy trade-offs across a wide range of configurations.

    \item We evaluate the proposed framework on an AlexNet-based brain tumour MRI classification model and compare it against standard low-rank baselines, including Global SVD and Tucker decomposition, under \(3\times\) and \(6\times\) compression budgets. The results, reported with bootstrap standard errors to quantify uncertainty, show that the proposed method achieves competitive performance at moderate compression and significantly outperforms baseline methods under more aggressive compression, while reducing FLOPs and inference latency.
\end{itemize}

The remainder of this paper is organised as follows. Section~\ref{sec:background} reviews background on low‑rank compression methods, with a focus on Global SVD and Tucker decomposition. Section~\ref{sec:method} presents the proposed hierarchical spatio‑channel compression mathematical framework in detail. Section~\ref{sec:performance_evaluation} discusses the evaluation metrics used to evaluate the performance of the models. Section~\ref{sec:experimental_setup} describes the experimental setup, and presents the empirical results along with a hyper‑parameter trade‑off analysis. Section~\ref{sec:discussion} discusses the findings and limitations. Finally, Section~\ref{sec:conclusion} concludes the paper and outlines directions for future work.

\section{Background}{\label{sec:background}}
\subsection{Global SVD for Convolutional Layers}
Consider a convolutional layer with weight tensor \(\mathcal{W} \in \mathbb{R}^{C_{\text{out}} \times C_{\text{in}} \times \kappa \times \kappa}\), where \(C_{\text{out}}\) and \(C_{\text{in}}\) denote the number of output and input channels, respectively, and \(\kappa\) is the spatial kernel size. To apply SVD, the weight tensor is reshaped into a matrix \(\mathbf{W} \in \mathbb{R}^{C_{\text{out}} \times (C_{\text{in}} \kappa^2)}\) by flattening the spatial and input channel dimensions. The SVD of \(\mathbf{W}\) is
\begin{align}
\mathbf{W} = \mathbf{U} \boldsymbol{\Sigma} \mathbf{V}^\top,
\end{align}
where \(\mathbf{U}\) and \(\mathbf{V}\) are orthogonal matrices and \(\boldsymbol{\Sigma}\) contains the singular values \(\sigma_1 \ge \sigma_2 \ge \dots \ge 0\). A rank‑\(r\) approximation retains the first \(r\) singular values and corresponding vectors
\begin{align}
\mathbf{W}_r = \mathbf{U}_r \boldsymbol{\Sigma}_r \mathbf{V}_r^\top\approx \mathbf{W},
\end{align}
where \(\mathbf{U}_r \in \mathbb{R}^{C_{\text{out}} \times r}\), \(\boldsymbol{\Sigma}_r \in \mathbb{R}^{r \times r}\), and \(\mathbf{V}_r \in \mathbb{R}^{(C_{\text{in}} \kappa^2) \times r}\). This approximation minimises the Frobenius norm error. In a compressed network, the layer is implemented as a basis convolution (using \(\mathbf{V}_r\) reshaped to \(r \times C_{\text{in}} \times \kappa \times \kappa\)) followed by a \(1\times1\) reconstruction convolution (using \(\mathbf{U}_r \boldsymbol{\Sigma}_r\)) \cite{denton2014exploiting}.

\subsection{Tucker Decomposition}
Unfolding a tensor along each mode enables the application of matrix SVD independently. This procedure forms the basis of the Tucker decomposition, which generalises the matrix SVD to tensors \cite{de2000multilinear}. For a third-order tensor \(\mathcal{X} \in \mathbb{R}^{I \times J \times K}\), the Tucker decomposition expresses it as a core tensor \(\mathcal{G} \in \mathbb{R}^{P \times Q \times R}\) multiplied by factor matrices along each mode:
\begin{align}
\mathcal{X} \approx \mathcal{G} \times_1 \mathbf{A} \times_2 \mathbf{B} \times_3 \mathbf{C},
\end{align}
where \(\mathbf{A} \in \mathbb{R}^{I \times P}\), \(\mathbf{B} \in \mathbb{R}^{J \times Q}\), and \(\mathbf{C} \in \mathbb{R}^{K \times R}\) are orthogonal factor matrices, and \(\times_n\) denotes the mode-\(n\) product. The multilinear rank of \(\mathcal{X}\) is the tuple \((P, Q, R)\), indicating the dimensionality along each mode after compression.

The Tucker decomposition is computed by performing matrix SVD on each mode-\(n\) unfolding of the tensor. For the mode-1 unfolding \(\mathbf{X}_{(1)} \in \mathbb{R}^{I \times JK}\), the left singular vectors give the factor matrix \(\mathbf{A}\). Similarly, \(\mathbf{B}\) and \(\mathbf{C}\) are obtained from the mode-2 and mode-3 unfoldings, respectively. The core tensor is then computed as
\begin{align}
\mathcal{G} = \mathcal{X} \times_1 \mathbf{A}^\top \times_2 \mathbf{B}^\top \times_3 \mathbf{C}^\top.
\end{align}
Truncating the factor matrices to \(P < I\), \(Q < J\), and \(R < K\) yields a low-multilinear-rank approximation of \(\mathcal{X}\), analogous to truncated SVD for matrices. 

In the context of neural network compression, the Tucker decomposition can be applied directly to the weight tensors of convolutional layers. For a 4D convolutional kernel \(\mathcal{W} \in \mathbb{R}^{C_{\text{out}} \times C_{\text{in}} \times \kappa \times \kappa}\), the Tucker‑2 variant factorises only the output and input channel modes, leaving the spatial dimensions unchanged
\begin{align}
\mathcal{W} \approx \mathcal{G} \times_1 \mathbf{U}^{(1)} \times_2 \mathbf{U}^{(2)},
\end{align}
where \(\mathcal{G} \in \mathbb{R}^{r_{\text{out}} \times r_{\text{in}} \times \kappa \times \kappa}\) is the core tensor, and \(\mathbf{U}^{(1)} \in \mathbb{R}^{C_{\text{out}} \times r_{\text{out}}}\) and \(\mathbf{U}^{(2)} \in \mathbb{R}^{C_{\text{in}} \times r_{\text{in}}}\) are factor matrices obtained from the mode-1 and mode-2 unfoldings, respectively. The compressed layer is implemented as three convolutions: a \(1\times1\) convolution for input channel reduction (using \(\mathbf{U}^{(2)}\)), a spatial convolution with \(\mathcal{G}\), and a \(1\times1\) convolution for output channel restoration (using \(\mathbf{U}^{(1)}\)) \cite{kim2015compression}. 

\section{Proposed Method}{\label{sec:method}}
\subsection{Method Overview}
\begin{figure*}
    \centering
    \includegraphics[width=1.0\linewidth]{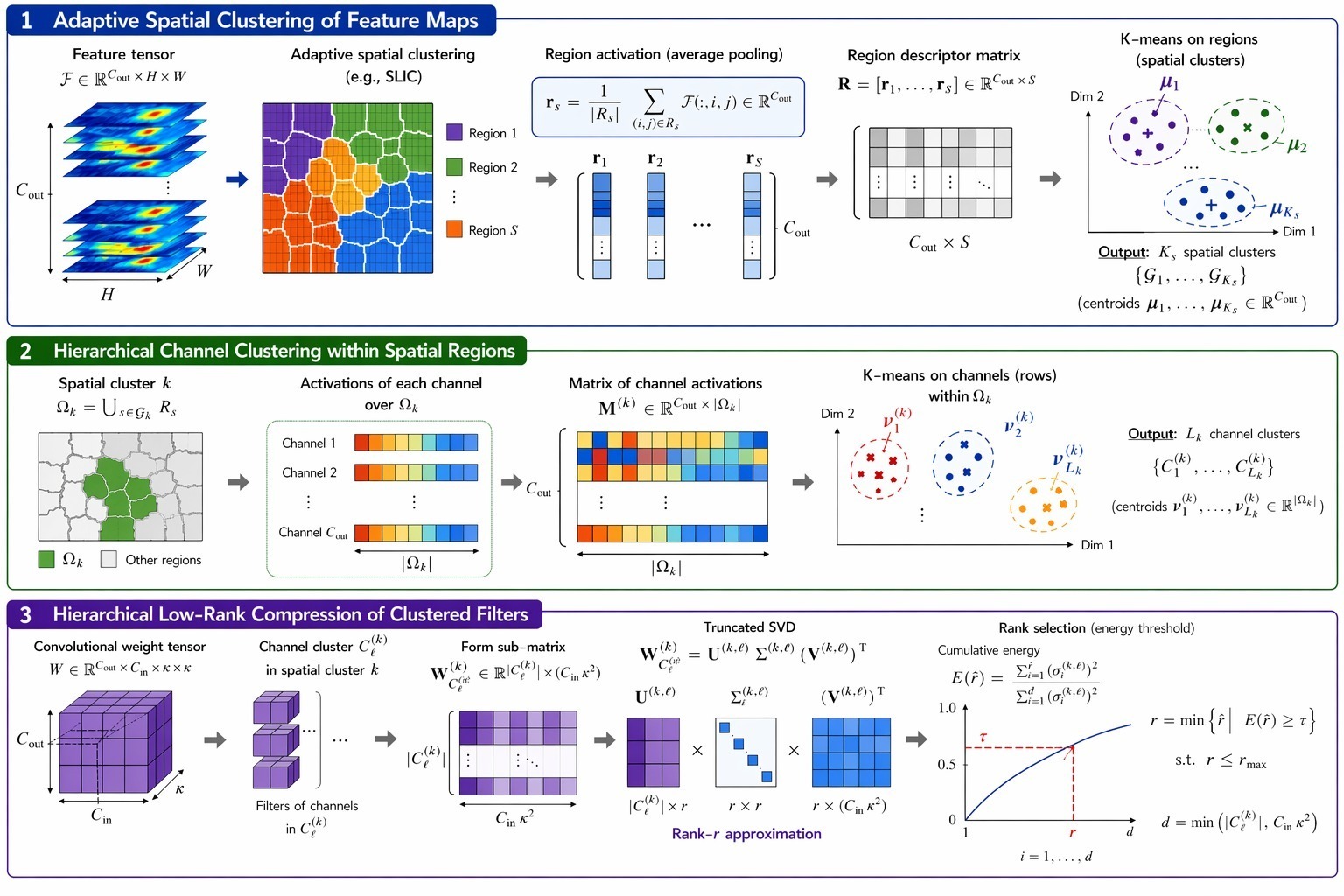}
    \caption{Overview of the proposed hierarchical spatio-channel low-rank compression framework.}
    \label{fig:method_overview}
\end{figure*}

An overview of the proposed framework is shown in Fig.~\ref{fig:method_overview}. The proposed method consists of three hierarchical steps. First, the output feature tensor of a convolutional layer is partitioned into spatially coherent regions using the simple linear iterative clustering (SLIC) algorithm. The resulting region descriptors are then grouped using K-means to identify spatial clusters that exhibit similar activation patterns across channels. Second, for each spatial cluster, the activations of all output channels over the corresponding spatial support are extracted and arranged into a data matrix. K-means is then applied to the rows of the data matrix to identify groups of channels that respond similarly within the selected spatial region. This produces a hierarchical decomposition in which spatial clustering captures where similar activation patterns occur, while channel clustering identifies which filters contribute to those patterns. Finally, the convolutional filters associated with each channel cluster are collected into a sub-matrix and compressed independently using truncated SVD. The approximation of the rank is selected adaptively using an energy threshold, retaining the smallest rank for which the cumulative sum of the retained singular values satisfies the prescribed proportion of the total energy. This allows each cluster to be compressed according to its intrinsic complexity, with more structured clusters retaining a higher rank and more redundant clusters admitting a lower-rank approximation.
\subsection{Adaptive Spatial Clustering of Feature Maps}
Spatial clustering is the process of identifying regions in feature maps that exhibit similar activation patterns across channels. Rather than treating each spatial location independently, the goal is to group nearby pixels whose activations form coherent structures. Given a feature tensor:
\begin{align}
\mathcal{F} \in \mathbb{R}^{C_{\text{out}} \times H \times W},
\end{align}
where \(H, W\) denotes the spatial dimensions. We seek to assign each pixel \((i,j)\) to one of \(S\) non-overlapping spatial regions. To accomplish this, we use a labelling function
\begin{align}
\ell: \{1,\dots,H\}\times\{1,\dots,W\} \to \{1,\dots,S\},
\end{align}
which partitions the spatial domain into disjoint regions \(R_1, \dots, R_S\), each capturing a localised pattern. That is, each region is defined as
\begin{align}
R_s = \{(i,j) \mid \ell(i,j)=s\},
\end{align}
and the collection \(\{R_s\}_{s=1}^S\) covers the entire grid, such that
\begin{equation}
\bigcup_{s=1}^S R_s = \{1,\dots,H\} \times \{1,\dots,W\}\,\text{and}\, R_s \cap R_{s'} = \emptyset \, \text{for } s \neq s'.
\end{equation}
In practice, \(\ell\) may be determined by an adaptive algorithm such as SLIC \cite{achanta2012slic}. It is important to note that SLIC is used solely to define spatially contiguous regions, while similarity between regions is determined later using channel-wise activation descriptors. Once regional assignments are set, each region \(R_s\) is summarized by a region activation vector \(\mathbf{r}_s \in \mathbb{R}^{C_{\text{out}}}\). Let
\begin{align}
\delta_s(i,j)=
\begin{cases}
1, & \ell(i,j)=s,\\
0, & \text{otherwise},
\end{cases}
\end{align}
denote the indicator function and,
\begin{align}
\vert R_s\vert=\sum_{i,j}\delta_s(i,j)
\end{align}
is defined as the number of pixels assigned to region \(s\), then the regional averages are given component-wise by
\begin{align}
[\mathbf{r}_s]_c 
= \frac{1}{\vert R_s\vert}\sum_{i=1}^H\sum_{j=1}^W 
\delta_s(i,j)\,\mathcal{F}_c(i,j),
\end{align}
or in vector form,
\begin{align}
\mathbf{r}_s 
= \frac{1}{\vert R_s\vert}\sum_{i,j} \delta_s(i,j)\,\mathcal{F}(:,i,j).
\end{align}
Stacking these vectors yields the region descriptor matrix
\begin{align}
\mathbf{R} = [\mathbf{r}_1,\mathbf{r}_2,\dots,\mathbf{r}_S]\in\mathbb{R}^{C_{\text{out}}\times S}.
\end{align}
To group similar regions, we cluster the columns of \(\mathbf{R}\) using K-means. K-means is suited for this step because the region descriptors \(\mathbf{r}_s\) lie in a Euclidean space, and the algorithm is computationally efficient, scaling linearly with the number of regions and the number of clusters, and clustering prior to low‑rank compression has been shown to be effective for medical images \cite{hamlomo2026clustering}. That is, K‑means identifies centroids \(\boldsymbol{\mu}_1,\dots,\boldsymbol{\mu}_{K_s}\in\mathbb{R}^{C_{\text{out}}}\) by minimizing
\begin{align}
\sum_{k=1}^{K_s}\sum_{s\in\mathcal{G}_k}
\|\mathbf{r}_s - \boldsymbol{\mu}_k\|_2^2,
\end{align}
where \(\{\mathcal{G}_1,\dots,\mathcal{G}_{K_s}\}\) forms a partition of \(\{1,\dots,S\}\). Through iterative assignment and update steps, the algorithm converges to clusters of regions whose average activation patterns across channels are most similar. Thus, by clustering these regional summaries, we obtain \(K_s\) spatial clusters that capture coherent patterns within the convolutional feature maps. 

Algorithm~\ref{alg:adaptive_spatial_clustering} performs adaptive clustering of spatial regions in the convolutional feature maps. Given the activation tensor \(\mathcal{F} \in \mathbb{R}^{C_{\mathrm{out}} \times H \times W}\), the algorithm first computes the mean activation at each spatial location by averaging across channels, yielding a 2D map. This map is then segmented into \(S\) super-pixels using SLIC. For each region \(R_s\), a descriptor vector \(\mathbf{r}_s \in \mathbb{R}^{C_{\mathrm{out}}}\) is obtained by averaging the activations over all pixels within the region. These vectors are stacked into the matrix \(\mathbf{R} \in \mathbb{R}^{C_{\mathrm{out}} \times S}\), whose columns summarise the regions. Finally, K-means clustering is applied to the columns of \(\mathbf{R}\), grouping similar regions into \(K_s\) spatial clusters \(\{\mathcal{G}_k\}\).

\begin{algorithm}[!htbp]
\caption{Adaptive Spatial Clustering via Superpixels}
\label{alg:adaptive_spatial_clustering}
\begin{algorithmic}[1]
\State \textbf{Input:} Feature map tensor \(\mathcal{F} \in \mathbb{R}^{C_{\text{out}} \times H \times W}\), number of superpixels \(S\), number of spatial clusters \(K_s\)
\State \textbf{Output:} Spatial clusters \(\{\mathcal{G}_1, \dots, \mathcal{G}_{K_s}\}\)
\State Compute mean activation map: \(\bar{\mathbf{F}}(i,j) \gets \frac{1}{C_{\text{out}}} \sum_{c=1}^{C_{\text{out}}} \mathcal{F}_{c}(i,j)\), for all \((i,j) \in \{1,\dots,H\} \times \{1,\dots,W\}\)
\State Segment \(\bar{\mathbf{F}}\) into \(S\) superpixels using SLIC: 
\begin{align}
\ell: \{1,\dots,H\}\times\{1,\dots,W\} \to \{1,\dots,S\}
\end{align}
\State Define superpixel regions: \(R_s \gets \{(i,j) \mid \ell(i,j) = s\}\), for \(s = 1,\dots,S\)
\State Initialize region descriptor matrix: \(\mathbf{R} \in \mathbb{R}^{C_{\text{out}} \times S}\)
\For{\(s \gets 1\) to \(S\)}
    \State Compute region average vector:
    \begin{align}
    \mathbf{r}_s \gets \frac{1}{\vert R_s \vert} \sum_{(i,j) \in R_s} \mathcal{F}(:, i, j)
    \end{align}
    \State Set column \(s\) of \(\mathbf{R}\): \([\mathbf{R}]_{:,s} \gets \mathbf{r}_s\)
\EndFor
\State Apply K-means to columns of \(\mathbf{R}\) to obtain region clusters:
\begin{align}
\{\mathcal{G}_1, \dots, \mathcal{G}_{K_s}\} \gets \text{KMeans}(\mathbf{R}, K_s)
\end{align}
\State \Return \(\{\mathcal{G}_1, \dots, \mathcal{G}_{K_s}\}\)
\end{algorithmic}
\end{algorithm}
\subsection{Hierarchical Channel Clustering within Spatial Regions}
Once spatial clusters \(\{\mathcal{G}_1, \dots, \mathcal{G}_{K_s}\}\) have been identified, the output channels are grouped according to their activation patterns within each region. For the \(k\)‑th spatial cluster, we have
\begin{align}
\Omega_k &= \bigcup_{s\in\mathcal{G}_k} R_s
\;\subseteq\;\{1,\dots,H\}\times\{1,\dots,W\}
\end{align}
with cardinality \(\vert \Omega_k\vert\). Extracting the activations of output channel \(c\) over this region yields
\begin{align}
\mathbf{f}_c^{(k)}
= \bigl[\mathcal{F}_c(i,j)\bigr]_{(i,j)\in\Omega_k}
\;\in\;\mathbb{R}^{\vert \Omega_k\vert},
\end{align}
and stacking these row‑wise for all \(c=1,\dots,C_{\text{out}}\) produces the matrix
\begin{align}
\mathbf{M}^{(k)}
&=
\begin{bmatrix}
(\mathbf{f}_1^{(k)})^\top\\
(\mathbf{f}_2^{(k)})^\top\\
\vdots\\
(\mathbf{f}_{C_{\text{out}}}^{(k)})^\top
\end{bmatrix}
\;\in\;\mathbb{R}^{C_{\text{out}}\times\vert \Omega_k\vert}.
\end{align}
To group channels whose activations exhibit similar behaviour over the spatial region \(\Omega_k\), we apply K-means clustering to the rows of \(\mathbf{M}^{(k)}\). Let \( L_k \) denote the number of channel clusters identified within the \( k \)-th spatial region. The goal is to find a partition
\begin{align}
\{\mathcal{C}_1^{(k)},\dots,\mathcal{C}_{L_k}^{(k)}\}
\end{align}
of the index set \(\{1,\dots,C_{\text{out}}\}\) that minimizes the within-cluster variance
\begin{align}
\sum_{\ell=1}^{L_k}
\sum_{c\in\mathcal{C}_\ell^{(k)}}
\bigl\|\mathbf{f}_c^{(k)} - \boldsymbol{\nu}_\ell^{(k)}\bigr\|_2^2,
\end{align}
where \(\boldsymbol{\nu}_\ell^{(k)} \in \mathbb{R}^{\vert \Omega_k \vert}\) denotes the centroid of cluster \(\mathcal{C}_\ell^{(k)}\), updated iteratively as
\begin{align}
\boldsymbol{\nu}_\ell^{(k)} = \frac{1}{\vert \mathcal{C}_\ell^{(k)} \vert} \sum_{c \in \mathcal{C}_\ell^{(k)}} \mathbf{f}_c^{(k)}.
\end{align}
Iterating assignment and update yields channel clusters whose activations co‑vary over \(\Omega_k\). Applying this independently for each \(k\) produces a hierarchical decomposition, that is, spatial clusters highlight where patterns arise, and channel clusters reveal which filters respond together.  

Algorithm~\ref{alg:channel_clustering} shows hierarchical channel clustering within each spatial region. For each spatial cluster \(\Omega_k\), it builds the matrix \(\mathbf{M}^{(k)}\in\mathbb{R}^{C_{\mathrm{out}}\times\vert\Omega_k\vert}\) whose rows are the flattened activation vectors \(\mathbf{f}_c^{(k)}\) of each output channel \(c\) over the pixels in \(\Omega_k\). It then applies K‑means to these row‑vectors to partition the set of channels into \(L_k\) groups \(\{\mathcal{C}_\ell^{(k)}\}\) whose activations exhibit correlated activation patterns within that region.
\begin{algorithm}[!htbp]
\caption{Channel Clustering within Spatial Clusters}
\label{alg:channel_clustering}
\begin{algorithmic}[1]
\State \textbf{Input:} Activation tensor \(\mathcal{F} \in \mathbb{R}^{C_{\text{out}} \times H \times W}\), spatial clusters \(\{\mathcal{G}_k\}_{k=1}^{K_s}\), number of channel clusters \(L_k\)
\State \textbf{Output:} Channel clusters \(\{\mathcal{C}_1^{(k)}, \dots, \mathcal{C}_{L_k}^{(k)}\}_{k=1}^{K_s}\)
\For{\(k \gets 1\) to \(K_s\)}
    \State Compute the spatial support of cluster \(k\):
    \begin{align}
    \Omega_k \gets \bigcup_{s\in\mathcal{G}_k} R_s
    \end{align}
    \State Initialize matrix \(\mathbf{M}^{(k)} \in \mathbb{R}^{C_{\text{out}} \times \vert \Omega_k \vert}\)
    \For{\(c \gets 1\) to \(C_{\text{out}}\)}
        \State Extract activation vector:
        \begin{align}
        \mathbf{f}_c^{(k)} \gets \left[\mathcal{F}_c(i,j)\right]_{(i,j)\in \Omega_k} \in \mathbb{R}^{\vert \Omega_k \vert}
        \end{align}
        \State Assign row \(c\) of \(\mathbf{M}^{(k)}\):
        \begin{align}
        \mathbf{M}^{(k)}[c,:] \gets \left( \mathbf{f}_c^{(k)} \right)^\top
        \end{align}
    \EndFor
    \State Perform K-means clustering over \(\{\mathbf{f}_c^{(k)}\}_{c=1}^{C_{\text{out}}}\) into \(L_k\) clusters
    \begin{align}
    \{\mathcal{C}_1^{(k)}, \dots, \mathcal{C}_{L_k}^{(k)}\} \gets \mathrm{KMeans}\!\left(\{\mathbf{f}_c^{(k)}\}_{c=1}^{C_{\text{out}}},\,L_k\right)
    \end{align} %\Comment{Each \(\mathcal{C}_\ell^{(k)} \subset \{1,\dots,C_{\text{out}}\}\) is a cluster of co-activating channels}
\EndFor
\State \Return \(\{\mathcal{C}_\ell^{(k)}\}_{\ell=1}^{L_k}, \quad \forall k \in \{1,\dots,K_s\}\)
\end{algorithmic}
\end{algorithm}
\subsection{Hierarchical Low‑Rank Compression of Clustered Filters}
To exploit both spatial and channel coherence, low‑rank approximation is applied independently to each channel cluster identified within every spatial region. Let \(\mathcal{C}_\ell^{(k)} \subset \{1,\dots,C_{\text{out}}\}\) denote the \(\ell\)-th output channel cluster within the \(k\)-th spatial region \(\Omega_k\). Let \(\mathcal{W} \in \mathbb{R}^{C_{\text{out}} \times C_{\text{in}} \times \kappa \times \kappa}\) denote the weight tensor of a convolutional layer, with \(C_{\text{out}}\) filters (output channels), each acting on \(C_{\text{in}}\) input channels with \(\kappa \times \kappa\) kernels. For each cluster \(\mathcal{C}_\ell^{(k)}\), we collect the filters corresponding to output channels in that cluster and flatten their spatial dimensions. Specifically, for each \(o\in\mathcal{C}_\ell^{(k)}\), the weight tensor slice \(\mathcal{W}_{o,:,:,:} \in \mathbb{R}^{C_{\text{in}} \times \kappa \times \kappa}\) is reshaped into a row vector in \(\mathbb{R}^{C_{\text{in}} \cdot \kappa^2}\). Stacking these gives
\begin{align}
\mathbf{W}^{(k)}_{\mathcal{C}_\ell}
\in \mathbb{R}^{\vert \mathcal{C}_\ell^{(k)} \vert \times (C_{\text{in}} \cdot \kappa^2)}.
\end{align}
This sub-matrix is expected to have a low-rank structure due to correlation among filters in the cluster. Although clustering is performed on activation patterns, filters that exhibit similar responses over spatial regions tend to encode correlated features. Therefore, grouping channels based on co-activation provides an implicit criterion for identifying redundancy in the corresponding convolutional filters, enabling effective low-rank approximation. For notational simplicity, let \(r \equiv r^{(k,\ell)}\) denote the selected rank for the current cluster \((k,\ell)\). Consequently, truncated SVD is applied to each sub-matrix, yielding the rank-\(r\) approximation
\begin{align}
\mathbf{W}^{(k)}_{\mathcal{C}_\ell,r}=\mathbf{U}_r^{(k,\ell)} \boldsymbol{\Sigma}_r^{(k,\ell)} \left( \mathbf{V}_r^{(k,\ell)} \right)^\top,
\end{align}
where \(\mathbf{U}_r^{(k,\ell)} \in \mathbb{R}^{\vert \mathcal{C}_\ell^{(k)} \vert \times r}\), 
\(\boldsymbol{\Sigma}_r^{(k,\ell)} \in \mathbb{R}^{r \times r}\), and 
\(\mathbf{V}_r^{(k,\ell)} \in \mathbb{R}^{(C_{\text{in}} \cdot \kappa^2) \times r}\)
denote the truncated left singular vectors, singular values, and right singular vectors, respectively. This rank-\(r\) approximation minimises the Frobenius norm error
\begin{align}
\left\Vert \mathbf{W}^{(k)}_{\mathcal{C}_\ell} - \mathbf{W}^{(k)}_{\mathcal{C}_\ell,r} \right\Vert_F^2
= \sum_{i = r + 1}^{\min\!\left(\vert\mathcal{C}_\ell^{(k)}\vert,\, C_{\mathrm{in}} \kappa^2\right)}
\bigl(\sigma_i^{(k,\ell)}\bigr)^2.
\end{align}
To adaptively select the rank \(r\), the energy threshold criterion is defined as
\begin{align}
r = \min \left\{ \hat{r} \;\middle\vert\;
\frac{\sum_{i=1}^{\hat{r}} \left( \sigma_i^{(k,\ell)} \right)^2}
{\sum_{i=1}^{\min\left( \left\lvert \mathcal{C}_\ell^{(k)} \right\rvert, C_{\text{in}} \cdot \kappa^2 \right)} \left( \sigma_i^{(k,\ell)} \right)^2}
\ge \tau \right\}\\
\text{such that} \quad r \le r_{\max},\nonumber
\end{align}
where $\tau \in (0,1]$ denotes the desired retained spectral energy and $r_{\max}$ denotes the maximum allowable rank to control computational cost.

Algorithm~\ref{alg:hierarchical_svd_compression} implements the adaptive SVD compression of the clustered filters (see \cite{hamlomo2026clustering} for an extensive discussion). For each channel cluster \(\mathcal{C}_\ell^{(k)}\), it extracts the corresponding rows of the weight tensor, flattening each \(\kappa\times\kappa\) kernel into a vector, to form the sub-matrix \(\mathbf{W}^{(k)}_{\mathcal{C}}\in\mathbb{R}^{\vert\mathcal{C}\vert\times(C_{\mathrm{in}}\kappa^2)}\). It then computes the singular values of this sub-matrix and determines the smallest rank \(r\) that preserves at least a fraction \(\tau\) of the total singular-value energy. Using this selected rank, the algorithm computes the corresponding truncated SVD factors \(\mathbf{U}_r^{(k,\ell)},\,\boldsymbol{\Sigma}_r^{(k,\ell)},\,\mathbf{V}_r^{(k,\ell)}\), which provide the optimal rank-\(r\) approximation of the clustered filter matrix in the Frobenius norm sense. These factors are later used to reconstruct each compressed convolutional layer as a basis, followed by a \(1\times1\) reconstruction convolution.

\begin{algorithm}[!htbp]
\caption{Hierarchical Low‑Rank SVD Compression of Clustered Filters}
\label{alg:hierarchical_svd_compression}
\begin{algorithmic}[1]
  \State \textbf{Input:} Weight tensor \(\mathcal{W}\in\mathbb{R}^{C_{\mathrm{out}}\times C_{\mathrm{in}}\times \kappa\times \kappa}\), channel clusters \(\{\mathcal{C}_\ell^{(k)}\}_{\ell=1,\dots,L_k}^{k=1,\dots,K_s}\), energy threshold \(\tau\in(0,1]\)
  \State \textbf{Output:} Truncated SVD factors \(\{\mathbf{U}_r^{(k,\ell)},\,\boldsymbol{\Sigma}_r^{(k,\ell)},\,\mathbf{V}_r^{(k,\ell)}\}\)

  \For{\(k = 1\) to \(K_s\)}
    \For{\(\ell = 1\) to \(L_k\)}
      \State Let \(\mathcal{C} = \mathcal{C}_\ell^{(k)}\), \(n = \lvert \mathcal{C}\rvert\)
      \If{\(n = 0\)}
        \State \textbf{continue} \Comment{Skip empty channel clusters}
      \EndIf
      \State Form sub-matrix:
      \begin{align}
        \mathbf{W}^{(k)}_{\mathcal{C}} =
        \begin{bmatrix}
          \mathrm{vec}(\mathcal{W}_{o,:,:,:})
        \end{bmatrix}_{o\in\mathcal{C}}
        \in\mathbb{R}^{n\times (C_{\mathrm{in}}\,\kappa^2)}
      \end{align}
      \State Compute the singular values of \(\mathbf{W}^{(k)}_{\mathcal{C}}\)
      \State Find smallest \(r\) such that
        \begin{align}
          \frac{\sum_{i=1}^{r} \sigma_i^2}
               {\sum_{i=1}^{\min(n,\,C_{\mathrm{in}}\,\kappa^2)} \sigma_i^2}
          \ge \tau
        \end{align}
      \State Compute the rank-\(r\) truncated SVD factors
        \begin{align}
          \mathbf{W}^{(k)}_{\mathcal{C}}
          \approx
          \mathbf{U}_r^{(k,\ell)}\,\boldsymbol{\Sigma}_r^{(k,\ell)}\,(\mathbf{V}_r^{(k,\ell)})^\top
        \end{align}
    \EndFor
  \EndFor
  \State \Return \(\{\mathbf{U}_r^{(k,\ell)},\,\boldsymbol{\Sigma}_r^{(k,\ell)},\,\mathbf{V}_r^{(k,\ell)}\}_{k=1,\ell=1}^{K_s,\,L_k}\)
\end{algorithmic}
\end{algorithm}
\subsection{Reconstruction of Compressed Weight Tensors}
To reconstruct the convolutional weight tensor from its low‑rank factors, the truncated SVD is inverted for each spatial region \(k\) and channel cluster \(\mathcal{C}_\ell^{(k)}\). Each row of the approximated weight matrix
\begin{align}
\mathbf{W}^{(k)}_{\mathcal{C}_\ell, r}
= \mathbf{U}_r^{(k,\ell)}\,\boldsymbol{\Sigma}_r^{(k,\ell)}\,\bigl(\mathbf{V}_r^{(k,\ell)}\bigr)^\top\in \mathbb{R}^{\vert \mathcal{C}_\ell^{(k)} \vert \times (C_{\text{in}} \cdot \kappa^2)}
\end{align}
represents a flattened version of a reconstructed convolutional kernel. For each output channel \( o_i \in \mathcal{C}_\ell^{(k)} \), the corresponding row \( \left[ \mathbf{W}_{\mathcal{C}_\ell, r}^{(k)} \right]_{i,:} \) is reshaped back to its original spatial dimensions through
\begin{align}
\widehat{\mathcal{W}}_{o_i,:,:,:}
= \operatorname{vec}^{-1}\Bigl( \bigl[\mathbf{W}^{(k)}_{\mathcal{C}_\ell, r} \bigr]_{i,:} \Bigr)
\in \mathbb{R}^{C_{\text{in}} \times \kappa \times \kappa},
\end{align}
for each output channel \(o_i \in \mathcal{C}_\ell^{(k)}\). Placing these reconstructed filters into their original positions yields the approximated weight tensor
\begin{align}
\widehat{\mathcal{W}}
= \bigl\{ \widehat{\mathcal{W}}_{o,:,:,:} \mid o = 1,\dots,C_{\text{out}} \bigr\}
\in \mathbb{R}^{C_{\text{out}} \times C_{\text{in}} \times \kappa \times \kappa}.
\end{align}
In addition to reconstructing the spatial weights, the bias vector from the original layer must also be preserved. Let \(\mathbf{b} \in \mathbb{R}^{C_{\text{out}}}\) denote the original bias vector. These are not affected by decomposition and are reused directly in the compressed layer. Each output feature map is thus reconstructed by applying the compressed convolution followed by the addition of the corresponding bias term, that is, 
\begin{align}
\mathcal{Y}_o = \widehat{\mathcal{W}}_{o,:,:,:} * \mathcal{X} + b_o,
\end{align} 
for each output channel \(o = 1, \dots, C_{\text{out}}\), where \(\mathcal{X} \in \mathbb{R}^{C_{\text{in}} \times H \times W}\) is the input tensor and \(*\) denotes the 2D convolution operation. The weights of the basis convolution are constructed from the reshaped right singular vectors \(\mathbf{V}_r^{(k,\ell)}\), while the reconstruction convolution encodes the product \(\mathbf{U}_r^{(k,\ell)} \boldsymbol{\Sigma}_r^{(k,\ell)}\), and the bias \(\mathbf{b}\) is assigned directly to the reconstruction layer to maintain the original affine structure of the network.

Algorithm~\ref{alg:reconstruction} reconstructs the full weight tensor by iterating over each region \(k\) and cluster \(\mathcal{C}_\ell^{(k)}\), applying \(\operatorname{vec}^{-1}\) to each row of \(\mathbf{W}^{(k)}_{\mathcal{C}_\ell, r}\) to recover its \(C_{\text{in}}\times \kappa\times \kappa\) kernel and placing it in \(\widehat{\mathcal{W}}\). Equivalently, the same compressed layer may be implemented directly as a basis convolution using kernels derived from \(\mathbf{V}_r^{(k,\ell)}\), followed by a \(1\times1\) reconstruction convolution parameterised by \(\mathbf{U}_r^{(k,\ell)}\boldsymbol{\Sigma}_r^{(k,\ell)}\).
\begin{algorithm}[!htbp]
\caption{Reconstruction of Compressed Weight Tensors}
\label{alg:reconstruction}
\begin{algorithmic}[1]
  \State \textbf{Input:} Compressed sub-matrices \(\{\mathbf{W}^{(k)}_{\mathcal{C}_\ell, r}\}_{k=1,\ell=1}^{K_s,\,L_k}\), channel clusters \(\{\mathcal{C}_\ell^{(k)}\}_{k=1,\ell=1}^{K_s,\,L_k}\), spatial kernel size \(\kappa\), dimensions \(C_{\text{out}}, C_{\text{in}}\)
  \State \textbf{Output:} Compressed weight tensor \(\widehat{\mathcal{W}}\)
  \State Initialize \(\widehat{\mathcal{W}}\gets \mathbf{0}\in \mathbb{R}^{C_{\text{out}}\times C_{\text{in}}\times \kappa\times \kappa}\)
  \For{\(k \gets 1\) to \(K_s\)}
    \For{\(\ell \gets 1\) to \(L_k\)}
      \State Let \(\mathcal{C} = \mathcal{C}_\ell^{(k)}\), \(n = \vert\mathcal{C}\vert\)
      \For{\(i \gets 1\) to \(n\)}
        \State \(o \gets \mathcal{C}[i]\)\Comment{output channel index}
        \State \(\mathbf{v} \gets \bigl[\mathbf{W}^{(k)}_{\mathcal{C}_\ell, r}\bigr]_{i,:}\)
        \State \(\widehat{\mathcal{W}}_{o,:,:,:} \gets \operatorname{vec}^{-1}(\mathbf{v})\)\Comment{reshape to \(C_{\text{in}}\!\times\!\kappa\!\times\!\kappa\)}
      \EndFor
    \EndFor
  \EndFor
  \State \Return \(\widehat{\mathcal{W}}\in\mathbb{R}^{C_{\text{out}}\times C_{\text{in}}\times \kappa\times \kappa}\)
\end{algorithmic}
\end{algorithm}

\subsection{Computational Complexity}
For a convolutional layer with \(C_{\mathrm{out}}\) output channels, \(C_{\mathrm{in}}\) input channels, kernel size \(\kappa\times\kappa\), and output feature map dimensions \(H\times W\), the additional computational cost introduced by the proposed framework arises from spatial clustering, channel clustering, and low-rank decomposition. The spatial clustering stage applies SLIC over the mean feature map. Since SLIC performs local \(K\)-means updates over a fixed neighbourhood of cluster centres, its complexity is linear in the number of pixels and is given by
\begin{align}
\mathcal{O}(HWI_s),
\end{align}
where \(I_s\) denotes the number of SLIC iterations \cite{achanta2012slic}. For each spatial cluster \(k\), channel clustering performs \(K\)-means over \(C_{\mathrm{out}}\) channel descriptors of dimension \(\vert\Omega_k\vert\). The computational complexity of Lloyd's \(K\)-means algorithm is linear in the number of data points, cluster centres, feature dimensions, and iterations \cite{lloyd1982least,arthur2007kmeans}. Therefore, the cost of channel clustering is
\begin{align}
\mathcal{O}\left(
\sum_{k=1}^{K_s}
C_{\mathrm{out}}\,L_k\,\vert\Omega_k\vert\,I_k
\right),
\end{align}
where \(I_k\) denotes the number of \(K\)-means iterations. For each channel cluster \(\mathcal{C}_\ell^{(k)}\), truncated SVD is applied to a matrix of size \(
\vert\mathcal{C}_\ell^{(k)}\vert\times(C_{\mathrm{in}}\kappa^2)\). Using an iterative or randomised truncated SVD algorithm, the complexity scales approximately linearly with the retained rank \(r\) \cite{halko2011finding,golub2013matrix}. For \(r \ll \min(\vert\mathcal{C}_\ell^{(k)}\vert, C_{\mathrm{in}}\kappa^2)\), the complexity is approximately
\begin{align}
\mathcal{O}\left(\sum_{k=1}^{K_s}\sum_{\ell=1}^{L_k}\vert\mathcal{C}_\ell^{(k)}\vert\,C_{\mathrm{in}}\kappa^2\,r\right).
\end{align}
The original convolution requires
\begin{align}
\mathcal{O}\left(
HWC_{\mathrm{out}}C_{\mathrm{in}}\kappa^2
\right)
\end{align}
operations. After compression, each cluster is implemented as a basis convolution followed by a \(1\times1\) reconstruction convolution, yielding a complexity of
\begin{align}
\mathcal{O}\left(
HW\sum_{k=1}^{K_s}\sum_{\ell=1}^{L_k}
\left[r\,C_{\mathrm{in}}\kappa^2 + \vert\mathcal{C}_\ell^{(k)}\vert\,r\right]\right).
\end{align}
Since \(r \ll \vert\mathcal{C}_\ell^{(k)}\vert\) and \(r \ll C_{\mathrm{in}}\kappa^2\), the compressed representation reduces both storage and computation relative to the original convolution.

\section{Performance Evaluation}{\label{sec:performance_evaluation}}
The performance of the compressed model $\widehat{f}(\mathbf{x}; \Theta)$ is evaluated on the test set 
\begin{align}
\mathcal{D}_{\text{test}} = \{(\mathbf{x}_i, y_i)\}_{i=1}^{N_{\text{test}}},
\end{align}
where $\Theta$ denotes the set of all parameters (weights and biases) of the compressed model $\widehat{f}(\mathbf{x}; \Theta)$ and each label $y_i \in \{1, \ldots, C_{\text{cls}}\}$ denotes one of the $C_{\text{cls}}$ output classes. For each input $\mathbf{x}_i$, the predicted class is defined by
\begin{align}
\widehat{y}_i = \arg\max_{c \in \{1, \ldots, C_{\text{cls}}\}} \widehat{f}_c(\mathbf{x}_i; \Theta),
\end{align}
where $\widehat{f}_c(\mathbf{x}_i; \Theta)$ denotes the predicted probability for class $c$. The overall classification accuracy is computed as the proportion of correctly predicted samples, that is
\begin{align}
\text{Accuracy} = \frac{1}{N_{\text{test}}} \sum_{i=1}^{N_{\text{test}}} \mathbb{I}[\widehat{y}_i = y_i],
\end{align}
with $\mathbb{I}[\cdot]$ the indicator function. Let $TP_c$, $FP_c$, and $FN_c$ denote the number of true positives, false positives, and false negatives, respectively, for class $c \in \{1, \ldots, C_{\text{cls}}\}$. The per-class recall and precision are then defined as
\begin{align}
\text{Precision}_c = \frac{TP_c}{TP_c + FP_c},
\end{align}
and 
\begin{align}
\text{Recall}_c = \frac{TP_c}{TP_c + FN_c}.
\end{align}
The harmonic mean of these two quantities yields the F1 score for class $c$, which is defined as 
\begin{align}
F1_c = \frac{2 \text{Precision}_c \times \text{Recall}_c}{\text{Precision}_c + \text{Recall}_c}.
\end{align}

To quantify the sampling variability of the point estimates derived from the finite test set, we employ the bootstrap \cite{efron1994introduction}. Let \(\hat{\theta}\) be a performance metric computed on \(\mathcal{D}_{\text{test}}\), such as accuracy, precision, recall, or the macro \(F_1\)-score. A total of \(B\) bootstrap samples
\begin{align}
\mathcal{D}_{\text{test}}^{*(1)}, \mathcal{D}_{\text{test}}^{*(2)}, \ldots, \mathcal{D}_{\text{test}}^{*(B)}
\end{align}
are generated by drawing \(N_{\text{test}}\) observations with replacement from \(\mathcal{D}_{\text{test}}\). For each bootstrap replicate \(b\), the metric is recomputed, yielding
\begin{align}
\hat{\theta}^{*(b)} = s\!\left(\mathcal{D}_{\text{test}}^{*(b)}\right),
\qquad b = 1, \dots, B,
\end{align}
where \(s(\cdot)\) denotes the statistic of interest. The bootstrap mean is
\begin{align}
\bar{\theta}^{*} = \frac{1}{B}\sum_{b=1}^{B}\hat{\theta}^{*(b)},
\end{align}
and the bootstrap standard error is given by
\begin{align}
\mathrm{SE}_{\mathrm{boot}}(\hat{\theta}) = \left[ \frac{1}{B-1}\sum_{b=1}^{B}\left( \hat{\theta}^{*(b)} - \bar{\theta}^{*} \right)^2 \right]^{1/2}.
\end{align}
This non‑parametric procedure requires no distributional assumptions and directly estimates the variability induced by the finite size of the test set.

The compression efficiency is quantified using three complementary metrics: \textit{parameter reduction}, \textit{FLOPs reduction}, and \textit{latency speed-up}. The original convolutional layer is parameterised by a weight tensor
\(\mathcal{W} \in \mathbb{R}^{C_{\mathrm{out}} \times C_{\mathrm{in}} \times \kappa \times \kappa}\),
so its total number of parameters is
\begin{align}
P_{\mathrm{orig}} = C_{\mathrm{out}} \times C_{\mathrm{in}} \times \kappa^2.
\end{align}
Bias parameters are excluded from this count for consistency, as they are unchanged during compression. After applying hierarchical clustering and SVD-based low-rank approximation, the layer is reparameterised in terms of compressed bases. For each spatial region 
\(k \in \{1, \ldots, K\}\) and channel cluster 
\(\ell \in \{1, \ldots, L_k\}\), the original sub-matrix
\begin{align}
\mathbf{W}_{C_\ell}^{(k)}\in \mathbb{R}^{n^{(k,\ell)} \times d},
\end{align}
where \(n^{(k,\ell)} = \vert C_\ell^{(k)} \vert\) and \(d = C_{\mathrm{in}} \times \kappa^2\) is approximated using a truncated SVD of rank \(r^{(k,\ell)}\). The number of parameters in the compressed representation becomes
\begin{align}
P_{\mathrm{comp}}^{(k,\ell)} = r^{(k,\ell)} \times \left(n^{(k,\ell)} + d\right).
\end{align}
This accounts for
\begin{align}
r^{(k,\ell)} \times d
\end{align}
parameters in the basis \(\mathbf{V}_r^{(k,\ell)}\) and
\begin{align}
r^{(k,\ell)} \times n^{(k,\ell)}
\end{align}
parameters in the projection and reconstruction matrix \(\mathbf{U}_r^{(k,\ell)} \boldsymbol{\Sigma}_r^{(k,\ell)}\). The per-cluster compression ratio is then defined as
\begin{align}
\mathrm{CR}^{(k,\ell)} = \frac{n^{(k,\ell)} \times d}{r^{(k,\ell)} \times \left(n^{(k,\ell)} + d\right)}.
\end{align}
This ratio captures the reduction in storage cost for the specific sub-matrix corresponding to cluster \((k,\ell)\). Therefore, the total number of parameters in the compressed layer \(l\) is given by 
\begin{align}
P_{\mathrm{comp}}^{(l)} = \sum_{k=1}^K \sum_{\ell=1}^{L_k} P_{\mathrm{comp}}^{(k,\ell)} 
= \sum_{k=1}^K \sum_{\ell=1}^{L_k} r^{(k,\ell)} \times \left(n^{(k,\ell)} + d\right).
\end{align}
The overall compression ratio for layer \(l\) is defined as
\begin{align}
\mathrm{CR}_{\mathrm{layer}} = \frac{P_{\mathrm{orig}}}{P_{\mathrm{comp}}^{(l)}}.
\end{align}
Since the compressed model replaces each original convolutional layer with its low-rank approximation, the overall model compression ratio is computed by summing parameters over all layers \(\mathcal{L}\), that is
\begin{align}
\mathrm{CR}_{\mathrm{model}} = \frac{\sum_{l \in \mathcal{L}} P_{\mathrm{orig}}^{(l)}}{\sum_{l \in \mathcal{L}} P_{\mathrm{comp}}^{(l)}}.
\end{align}
The percentage reduction in parameters is then given by
\begin{align}
\Delta P = \left(1 - \frac{\sum_{l \in \mathcal{L}} P_{\mathrm{comp}}^{(l)}}{\sum_{l \in \mathcal{L}} P_{\mathrm{orig}}^{(l)}}\right) \times 100\%.
\end{align}

In addition to parameter reduction, we evaluate the reduction in floating-point operations (FLOPs). The clustering techniques (spatial and channel-wise) are performed offline during the compression stage and do not contribute to inference-time computational cost. Therefore, all reported FLOPs and latency measurements reflect only the forward pass of the compressed model. Following the multiply-accumulate convention \cite{molchanov2017pruning, he2017channel}, the original layer's FLOPs are
\begin{align}
\text{FLOPs}_{\mathrm{orig}} = H' \times W' \times C_{\mathrm{out}} \times C_{\mathrm{in}} \times \kappa^2,
\end{align}
where \(H' \times W'\) are the output spatial dimensions. After compression, each original convolution is replaced by a basis convolution (using the right singular vectors) and a \(1 \times 1\) projection convolution. For a given spatial region \(k\) and channel cluster \(\mathcal{C}_\ell^{(k)}\) with rank \(r^{(k,\ell)}\), the FLOPs for the basis convolution are
\begin{align}
\text{FLOPs}_{\mathrm{basis}}^{(k,\ell)} = H' \times W' \times r^{(k,\ell)} \times C_{\mathrm{in}} \times \kappa^2,
\end{align}
and for the projection convolution are
\begin{align}
\text{FLOPs}_{\mathrm{proj}}^{(k,\ell)} = H' \times W' \times \vert \mathcal{C}_\ell^{(k)} \vert \times r^{(k,\ell)}.
\end{align}
Summing over all clusters and spatial regions gives the total FLOPs for the compressed layer
\begin{align}
\text{FLOPs}_{\mathrm{comp}}^{(l)}
&= H' \times W' \times \sum_{k=1}^K \sum_{\ell=1}^{L_k}
\Bigl(
r^{(k,\ell)} \times C_{\mathrm{in}} \times \kappa^2
\nonumber\\
&\qquad\qquad
+ \vert \mathcal{C}_\ell^{(k)} \vert \times r^{(k,\ell)}
\Bigr).
\end{align}
The FLOPs reduction ratio for the model is then
\begin{align}
\mathrm{FLOPs}_{\mathrm{ratio}} = \frac{\sum_{l \in \mathcal{L}} \text{FLOPs}_{\mathrm{orig}}^{(l)}}{\sum_{l \in \mathcal{L}} \text{FLOPs}_{\mathrm{comp}}^{(l)}},
\end{align}
and the percentage reduction is
\begin{align}
\Delta \text{FLOPs} = \left(1 - \frac{\sum_{l \in \mathcal{L}} \text{FLOPs}_{\mathrm{comp}}^{(l)}}{\sum_{l \in \mathcal{L}} \text{FLOPs}_{\mathrm{orig}}^{(l)}}\right) \times 100\%.
\end{align}

Finally, we measure the empirical latency speed-up on the target hardware. Let \(T_{\mathrm{orig}}\) and \(T_{\mathrm{comp}}\) denote the average inference times of the original and compressed models, respectively, measured under identical conditions. The latency speed-up factor is
\begin{align}
\text{speed-up} = \frac{T_{\mathrm{orig}}}{T_{\mathrm{comp}}},
\end{align}
and the percentage reduction in latency is
\begin{align}
\Delta T = \left(1 - \frac{T_{\mathrm{comp}}}{T_{\mathrm{orig}}}\right) \times 100\%.
\end{align}

\section{Experimental Setup}{\label{sec:experimental_setup}}
We train an AlexNet-based classifier \cite{krizhevsky2012imagenet} as the baseline model. The architecture consists of five convolutional layers (\texttt{Conv1}-\texttt{Conv5}) followed by three fully connected layers, taking \(224\times224\) RGB inputs. We then apply three compression methods: Global SVD, Tucker-2, and our proposed hierarchical spatio-channel compression method. Global SVD and Tucker-2 serve as baselines for comparison. 

Specifically, Global SVD reshapes each convolutional weight tensor into a matrix and factorised using SVD, following the approach of Denton et al.~\cite{denton2014exploiting}. The rank is chosen for each layer to minimise the difference between the resulting parameter count and a target budget derived from the overall compression ratio. The compressed layer is then reparameterised as a basis convolution followed by a \(1\times1\) reconstruction convolution. On the other hand, Tucker‑2 decomposition factorises the weight tensor into a core tensor and two factor matrices. The ranks are selected by a grid search over candidate values to approximate the target parameter budget, as described by Kim et al.~\cite{kim2015compression}. The decomposition is implemented as a sequence of three convolutions: a \(1\times1\) input projection, a \(\kappa\times\kappa\) core convolution, and a \(1\times1\) output projection. Both baselines are applied to the same set of convolutional layers and are fine‑tuned under identical conditions.

The training procedure for the baseline model and all compressed variants uses stochastic gradient descent with momentum 0.9, weight decay 1e‑4, and an initial learning rate of 0.1, annealed with a cosine schedule. All models are fine‑tuned for 30 epochs using stochastic gradient descent with momentum 0.9, weight decay 1e‑4, a learning rate of 1e‑3, and a cosine annealing schedule with a three‑epoch warm‑up. Early stopping is applied after five epochs without improvement on the validation set.

\subsection{Dataset Description}
The Brain Tumour Classification (MRI) dataset \cite{bhuvaji2020brain}, publicly available on Kaggle, provides T1-weighted magnetic resonance imaging (MRI) scans for multi-class brain tumour classification. The dataset comprises grayscale MRI images organised into four clinically relevant categories: glioma tumour, meningioma tumour, pituitary tumour, and no tumour. Gliomas originate from glial cells, meningiomas arise from the meninges, and pituitary tumours involve the pituitary gland, while the no tumour class includes scans without any visible pathology. The dataset contains approximately 3,264 images, distributed unevenly across classes, and is structured into separate training and test directories, each containing subfolders corresponding to the four tumour categories. The images vary in resolution and anatomical view, including sagittal, axial, and coronal planes. Training set images were initially resized to 256×256 pixels, followed by data augmentation including random resized cropping to 224×224 pixels, horizontal flipping, small-angle rotations, affine translations and scaling, and minor brightness and contrast variations. These augmentations were introduced to improve model generalisation by increasing variability in the training data while preserving the underlying anatomical structure. Since the original MRI scans are grayscale, each image was converted to a three-channel representation to ensure compatibility with CNN architectures pretrained on natural image datasets. All images were subsequently normalised using the standard ImageNet mean and standard deviation \cite{krizhevsky2012imagenet}.

\begin{figure}[ht]
    \centering
    \includegraphics[width=1.0\linewidth]{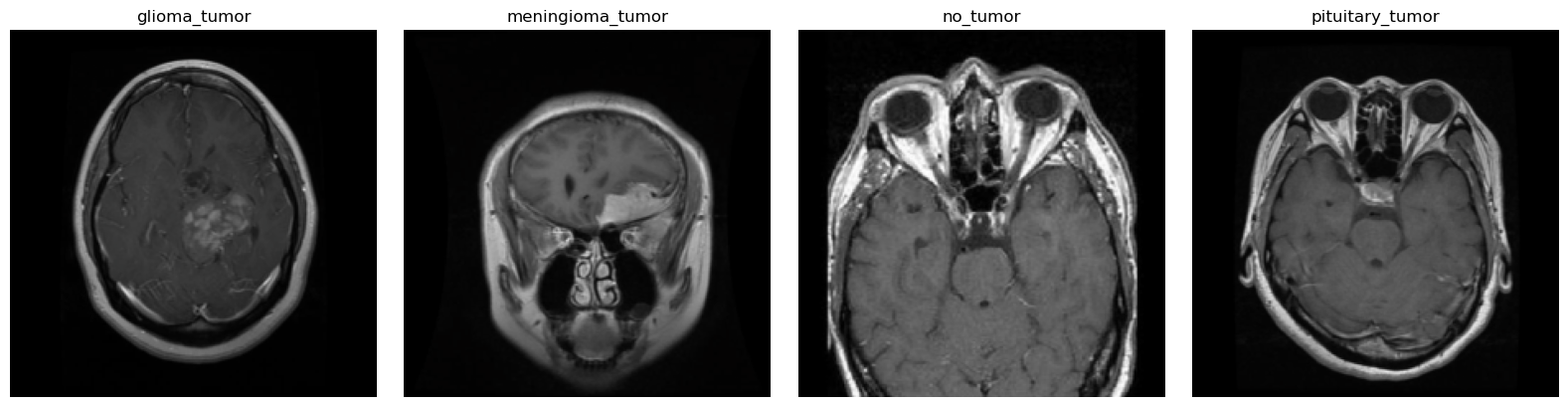}
    \caption{Representative T1-weighted MRI slices from the training set showing one sample from each class: glioma tumour, meningioma tumour, no tumour, and pituitary tumour. These samples highlight differences in tumour location, shape, and intensity patterns, as well as the absence of pathological structures in the no tumour class.}
    \label{fig:mri_data}
\end{figure}

The original training set was further split into training and validation subsets at a 80/20 ratio. To address class imbalance, class-dependent weights inversely proportional to the number of samples in each class were computed and incorporated into a weighted sampling scheme during training. Representative samples from each class in the training set are shown in Fig.~\ref{fig:mri_data}, illustrating the variability in anatomical structure, tumour appearance, and intensity characteristics across the different categories.

\subsection{Performance Evaluation}
We evaluate the proposed adaptive spatio-channel low-rank compression framework on the brain tumour MRI classification dataset by analysing layer-wise compression behaviour, computational efficiency, and classification performance. The aim is not only to compare the final classification accuracy of the different compression budgets, but also to examine how each method distributes its compression budget across the individual convolutional layers. All classification metrics are reported with bootstrap standard errors computed using \(B = 1000\) resamples of the test set predictions, as described in Section~\ref{sec:performance_evaluation}. This provides a direct estimate of the sampling variability due to the finite test set size.

Tables~\ref{tab:layer_3x} and~\ref{tab:layer_6x} show the parameter reduction achieved in each compressed layer. The purpose of this analysis is to determine whether the methods compress all layers uniformly or whether they allocate the compression budget differently across shallow and deep convolutional layers.

At the \(3\times\) budget, both Global SVD and Tucker decomposition exhibit an almost uniform reduction pattern, compressing every layer by approximately \(66\text{--}67\%\). This is expected because both methods apply low‑rank approximation to each layer independently under a Global parameter budget, without explicit guidance from layer‑specific activation redundancy. Consequently, they tend to distribute the compression budget nearly uniformly across layers \cite{idelbayev2020low}. In contrast, the proposed method distributes the compression budget non-uniformly. \texttt{Conv2} is compressed much more strongly, by \(74.9\%\), than \texttt{Conv3}, \texttt{Conv4}, and \texttt{Conv5}, whose reductions remain between \(64.8\%\) and \(67.7\%\). This indicates that the proposed activation-guided clustering identified substantially more redundancy in the early convolutional filters than in the deeper layers. 

\begin{table}
\centering
\caption{Per-layer parameter reduction (\%) at the \(3\times\) compression budget.}
\label{tab:layer_3x}
\setlength{\tabcolsep}{7pt}
\begin{tabular}{lcccc}
\toprule
Method & \texttt{Conv2} & \texttt{Conv3} & \texttt{Conv4} & \texttt{Conv5} \\
\midrule
Global SVD & 66.6 & 66.5 & 66.8 & 66.6 \\
Tucker     & 66.7 & 66.6 & 67.4 & 66.9 \\
Our Method & 74.9 & 65.6 & 67.7 & 64.8 \\
\bottomrule
\end{tabular}
\end{table}

At the more aggressive \(6\times\) budget, the same trend becomes more evident. Global SVD and Tucker again distribute the compression almost uniformly, reducing all layers by approximately \(83\text{--}84\%\). The proposed method instead shifts the compression budget toward the deeper layers. \texttt{Conv2} is now compressed less aggressively, by \(75.6\%\), than the other layers, while \texttt{Conv4} receives the strongest compression at \(84.5\%\). Thus, as the overall budget becomes tighter, the proposed method retains more parameters in the earlier convolutional layer while exploiting greater redundancy in the deeper layers. This suggests that the earlier layer contains more task-critical low-level features that are sensitive to over-compression, whereas the deeper layers contain more redundant filters that can be compressed more aggressively.

\begin{table}
\centering
\caption{Per-layer parameter reduction (\%) at the \(6\times\) compression budget.}
\label{tab:layer_6x}
\setlength{\tabcolsep}{7pt}
\begin{tabular}{lcccc}
\toprule
Method & \texttt{Conv2} & \texttt{Conv3} & \texttt{Conv4} & \texttt{Conv5} \\
\midrule
Global SVD & 83.2 & 83.4 & 83.2 & 83.5 \\
Tucker     & 83.0 & 82.8 & 84.2 & 82.9 \\
Our Method & 75.6 & 83.4 & 84.5 & 83.0 \\
\bottomrule
\end{tabular}
\end{table}

Table~\ref{tab:overall_3x} summarises the overall model performance at the \(3\times\) compression budget. The proposed method achieves the highest accuracy, \(87.96\%\), slightly improving upon the uncompressed baseline at \(87.76\%\) and Global SVD at \(87.76\%\), while substantially outperforming Tucker decomposition at \(84.69\%\). Although all compressed models reduce the total number of parameters by \(5.3\%\), because the fully connected layers dominate the parameter count, the reduction in convolutional computation is much more substantial. The proposed method reduces FLOPs from 8.21\,G to 2.73\,G and decreases latency from 1.80\,ms to 1.44\,ms, corresponding to a \(1.26\times\) speed-up.
\begin{table*}
\centering
\caption{Overall performance comparison at $3\times$ compression. Accuracy is reported as mean $\pm$ standard error (\%). Best value in each column is shown in bold.}
\label{tab:overall_3x}
\resizebox{\textwidth}{!}{%
\begin{tabular}{lcccccc}
\toprule
Method & Acc. (\%) & Params (M) & Size (MB) & FLOPs (G) & Latency (ms) & Speed-up \\
\midrule
Baseline   & 87.76 $\pm$ 1.52 & 35.83 & 136.68 & 8.21 & 1.80 & 1.00$\times$ \\
Global SVD & 87.76 $\pm$ 1.51 & 34.33 & 130.96 & 2.82 & 1.49 & 1.21$\times$ \\
Tucker     & 84.69 $\pm$ 1.60 & \textbf{34.32} & \textbf{130.93} & 2.79 & \textbf{1.43} & \textbf{1.26$\times$} \\
Our Method & \textbf{87.96 $\pm$ 1.44} & 34.33 & 130.96 & \textbf{2.73} & 1.44 & \textbf{1.26$\times$} \\
\bottomrule
\end{tabular}%
}
\end{table*}

Table~\ref{tab:report_3x} shows per‑class precision, recall, and \(F_1\)-score at the \(3\times\) compression budget. The proposed method achieves the highest \(F_1\)-score for the pituitary class, \(0.945\), and the second-highest for the meningioma class, \(0.844\), while its performance on the no‑tumour class, \(0.872\), is comparable to the best results. Global SVD attains the best \(F_1\)-score for the glioma class, \(0.873\), and the no‑tumour class, \(0.892\). Its meningioma \(F_1\)-score, \(0.839\), falls slightly below the baseline value of \(0.848\). Tucker decomposition consistently yields the lowest metrics across all classes, with the largest declines observed for meningioma and pituitary tumours. At the macro level, the proposed method attains a recall of \(0.883\), slightly below Global SVD's \(0.884\), while maintaining precision close to that of the baseline, indicating that compression does not substantially compromise the model's ability to identify individual classes.

\begin{table}
\centering
\caption{Class-wise precision, recall, and $F_1$-score at the moderate $3\times$ compression budget. Values are reported as mean $\pm$ standard error (\%). For each metric row, the highest value among all methods is highlighted in bold.}
\label{tab:report_3x}
\footnotesize
\setlength{\tabcolsep}{3pt}
\begin{tabular}{@{}lcccccc@{}}
\toprule
Class & Metric & Baseline & Global SVD & Tucker & Our Method \\
\midrule
Glioma   & Precision & 85.2 $\pm$ 3.0 & 88.2 $\pm$ 2.8 & 75.2 $\pm$ 3.3 & \textbf{91.1 $\pm$ 2.6} \\
         & Recall    & 87.1 $\pm$ 2.9 & 86.3 $\pm$ 2.9 & \textbf{91.4 $\pm$ 2.4} & 80.6 $\pm$ 3.4 \\
         & $F_1$      & 86.1 $\pm$ 2.2 & \textbf{87.3 $\pm$ 2.2} & 82.5 $\pm$ 2.3 & 85.5 $\pm$ 2.4 \\
\midrule
Meningioma & Precision & 86.7 $\pm$ 3.1 & 84.8 $\pm$ 3.1 & \textbf{92.7 $\pm$ 2.4} & 82.4 $\pm$ 3.1 \\
           & Recall    & 83.0 $\pm$ 3.1 & 83.0 $\pm$ 3.1 & 71.6 $\pm$ 3.7 & \textbf{86.5 $\pm$ 2.8} \\
           & $F_1$      & \textbf{84.8 $\pm$ 2.3} & 83.9 $\pm$ 2.4 & 80.8 $\pm$ 2.6 & 84.4 $\pm$ 2.2 \\
\midrule
No tumour & Precision & \textbf{87.2 $\pm$ 3.9} & 85.4 $\pm$ 3.9 & 81.4 $\pm$ 4.2 & 84.0 $\pm$ 4.2 \\
          & Recall    & 90.7 $\pm$ 3.4 & \textbf{93.3 $\pm$ 2.9} & \textbf{93.3 $\pm$ 2.9} & 90.7 $\pm$ 3.4 \\
          & $F_1$      & 88.9 $\pm$ 2.7 & \textbf{89.2 $\pm$ 2.6} & 87.0 $\pm$ 2.8 & 87.2 $\pm$ 2.9 \\
\midrule
Pituitary & Precision & 91.9 $\pm$ 2.4 & 91.8 $\pm$ 2.4 & 92.9 $\pm$ 2.4 & \textbf{93.5 $\pm$ 2.1} \\
          & Recall    & 91.9 $\pm$ 2.5 & 91.1 $\pm$ 2.5 & 86.7 $\pm$ 3.0 & \textbf{95.6 $\pm$ 1.8} \\
          & $F_1$      & 91.9 $\pm$ 1.8 & 91.5 $\pm$ 1.8 & 89.7 $\pm$ 2.0 & \textbf{94.5 $\pm$ 1.5} \\
\midrule
Macro avg & Precision & 87.7 $\pm$ 1.6 & 87.5 $\pm$ 1.6 & 85.5 $\pm$ 1.6 & \textbf{87.8 $\pm$ 1.6} \\
          & Recall    & 88.1 $\pm$ 1.5 & \textbf{88.4 $\pm$ 1.5} & 85.8 $\pm$ 1.5 & 88.3 $\pm$ 1.5 \\
          & $F_1$      & \textbf{87.9 $\pm$ 1.5} & \textbf{87.9 $\pm$ 1.5} & 85.0 $\pm$ 1.6 & \textbf{87.9 $\pm$ 1.5} \\
\midrule
Weighted avg & Precision & 87.8 $\pm$ 1.5 & 87.8 $\pm$ 1.5 & 86.0 $\pm$ 1.4 & \textbf{88.2 $\pm$ 1.4} \\
             & Recall    & 87.8 $\pm$ 1.5 & 87.8 $\pm$ 1.5 & 84.7 $\pm$ 1.6 & \textbf{88.0 $\pm$ 1.4} \\
             & $F_1$      & 87.7 $\pm$ 1.5 & 87.7 $\pm$ 1.5 & 84.7 $\pm$ 1.6 & \textbf{87.9 $\pm$ 1.5} \\
\bottomrule
\end{tabular}
\normalsize
\end{table}

\begin{figure*}
    \centering
    \includegraphics[width=1.0\linewidth]{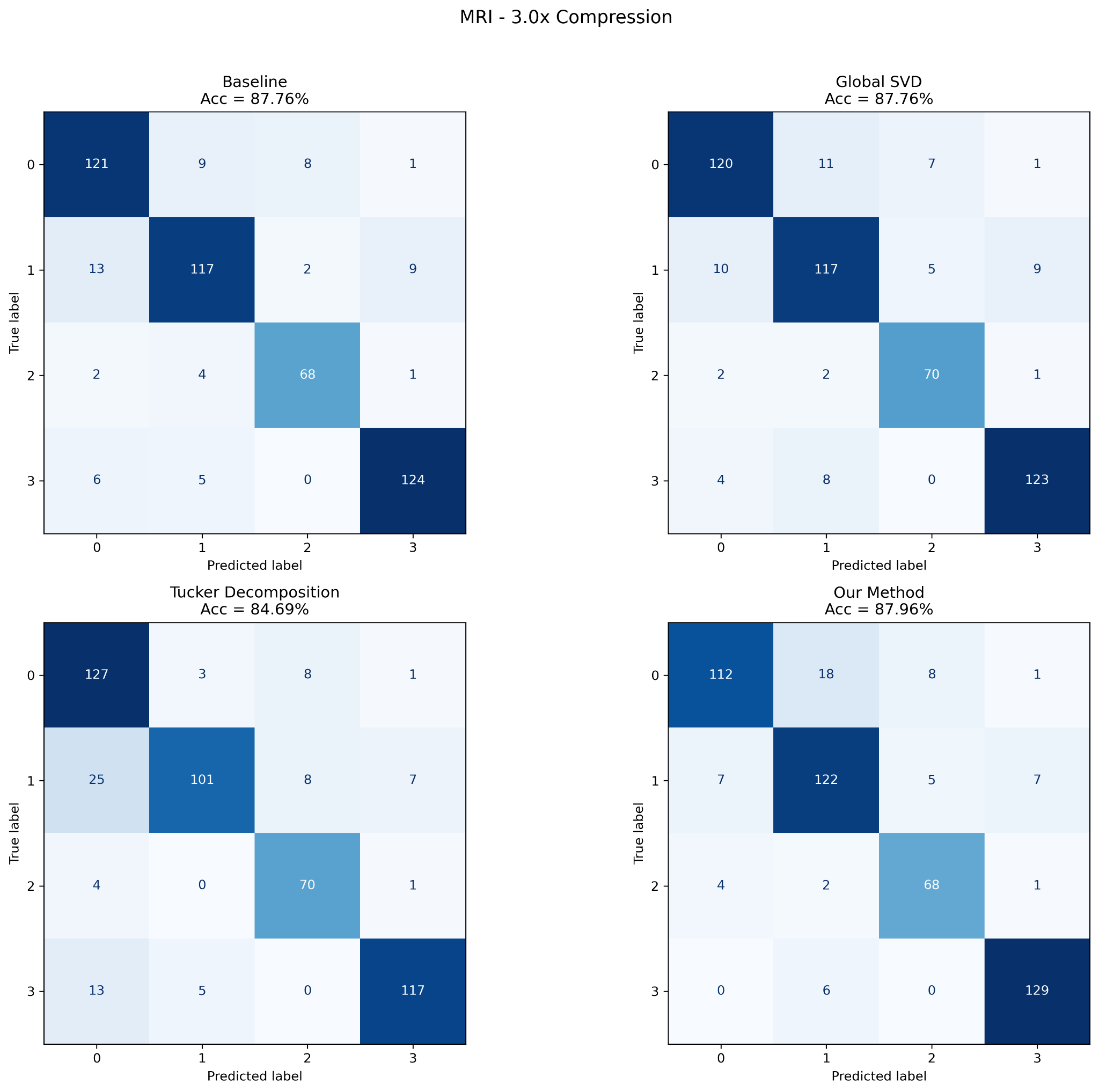}
    \caption{Confusion matrices for the baseline model, Global SVD, Tucker decomposition, and the proposed method at the \(3\times\) compression budget. The proposed method remains closest to the original classifier while increasing the number of correctly classified meningioma and pituitary cases.}
    \label{fig:confusion_3x}
\end{figure*}

The confusion matrices in Fig.~\ref{fig:confusion_3x} confirm these trends. At the \(3\times\) budget, all methods preserve much of the baseline structure, but Tucker decomposition introduces noticeably more confusion between glioma and meningioma. The proposed method remains much closer to the original model while correctly classifying more meningioma and pituitary samples. In particular, the number of correctly classified meningioma cases increases from \(117\) to \(122\), while the number of correctly classified pituitary cases increases from \(124\) to \(129\).

At the \(6\times\) budget, shown in Table~\ref{tab:overall_6x}, the advantage of the proposed method becomes much more evident. The proposed method attains \(89.80\%\) accuracy, exceeding the baseline by 2.04\% and substantially outperforming both Global SVD, \(84.90\%\), and Tucker decomposition, \(85.31\%\). This may be because, at high compression, the method acts as a form of feature extraction by filtering out background noise in the data. The model size decreases to 129.53\,MB and FLOPs fall to 1.55\,G, corresponding to an \(81.1\%\) reduction relative to the original model. Latency decreases to 1.30\,ms, producing a \(1.38\times\) speed‑up. Notably, Tucker decomposition achieves the lowest latency, \(1.11\)\,ms, and highest speed‑up, \(1.62\times\), at this budget, although its accuracy drops to \(85.31\%\), whereas the proposed method attains \(89.80\%\) accuracy with a speed‑up of \(1.38\times\).

\begin{table*}
\centering
\caption{Overall performance comparison at $6\times$ compression. Accuracy is reported as mean $\pm$ standard error (\%). Best value in each column is shown in bold.}
\label{tab:overall_6x}
\resizebox{\textwidth}{!}{%
\begin{tabular}{lcccccc}
\toprule
Method & Acc. (\%) & Params (M) & Size (MB) & FLOPs (G) & Latency (ms) & Speed-up \\
\midrule
Baseline   & 87.76 $\pm$ 1.52 & 35.83 & 136.68 & 8.21 & 1.80 & 1.00$\times$ \\
Global SVD & 84.90 $\pm$ 1.63 & 33.96 & 129.53 & 1.47 & 1.21 & 1.49$\times$ \\
Tucker     & 85.31 $\pm$ 1.59 & \textbf{33.95} & \textbf{129.52} & \textbf{1.46} & \textbf{1.11} & \textbf{1.62$\times$} \\
Our Method & \textbf{89.80 $\pm$ 1.37} & 33.96 & 129.53 & 1.55 & 1.30 & 1.38$\times$ \\
\bottomrule
\end{tabular}%
}
\end{table*}

Table~\ref{tab:report_6x} reports per‑class precision, recall, and \(F_1\)-score at the \(6\times\) compression budget. The proposed method achieves the highest \(F_1\)-score for the meningioma, no‑tumour, and pituitary classes: \(0.879\), \(0.883\), and \(0.949\), respectively. For the glioma class, the proposed method attains the best \(F_1\)-score, \(0.875\), followed closely by the baseline at \(0.861\). Global SVD and Tucker decomposition exhibit declines across all classes, with macro \(F_1\)-scores of \(0.849\) and \(0.856\), respectively, compared to \(0.879\) for the baseline and \(0.896\) for the proposed method. The largest gains for the proposed method appear in the meningioma class, where the \(F_1\)-score improves from \(0.848\) (baseline) to \(0.879\), and in the pituitary class, where recall increases from \(0.919\) to \(0.963\). At the macro level, the proposed method outperforms the baseline by \(1.7\%\) in precision, \(1.9\%\) in recall, and \(1.7\%\) in \(F_1\)-score, suggesting that structured compression preserves, and in some cases improves, class‑wise discrimination under high compression.

\begin{table}
\centering
\caption{Class-wise precision, recall, and $F_1$-score under the aggressive $6\times$ compression budget. Values are reported as mean $\pm$ standard error (\%). Bold entries denote the best performance per row across the baseline and compressed models.}
\label{tab:report_6x}
\footnotesize
\setlength{\tabcolsep}{3pt}
\begin{tabular}{@{}lcccccc@{}}
\toprule
Class & Metric & Baseline & Global SVD & Tucker & Our Method \\
\midrule
Glioma   & Precision & 85.2 $\pm$ 3.0 & 83.9 $\pm$ 3.2 & 86.5 $\pm$ 3.0 & \textbf{89.5 $\pm$ 2.7} \\
         & Recall    & \textbf{87.1 $\pm$ 2.9} & 82.7 $\pm$ 3.2 & 82.7 $\pm$ 3.2 & 85.6 $\pm$ 2.9 \\
         & $F_1$      & 86.1 $\pm$ 2.2 & 83.3 $\pm$ 2.5 & 84.6 $\pm$ 2.4 & \textbf{87.5 $\pm$ 2.1} \\
\midrule
Meningioma & Precision & 86.7 $\pm$ 3.1 & 82.1 $\pm$ 3.3 & 76.1 $\pm$ 3.5 & \textbf{88.5 $\pm$ 2.7} \\
           & Recall    & 83.0 $\pm$ 3.1 & 78.0 $\pm$ 3.4 & 85.8 $\pm$ 2.8 & \textbf{87.2 $\pm$ 2.7} \\
           & $F_1$      & 84.8 $\pm$ 2.3 & 80.0 $\pm$ 2.6 & 80.7 $\pm$ 2.5 & \textbf{87.9 $\pm$ 2.0} \\
\midrule
No tumour & Precision & 87.2 $\pm$ 3.9 & 80.7 $\pm$ 4.4 & \textbf{88.7 $\pm$ 3.8} & 86.1 $\pm$ 3.9 \\
          & Recall    & \textbf{90.7 $\pm$ 3.4} & 89.3 $\pm$ 3.5 & 84.0 $\pm$ 4.2 & \textbf{90.7 $\pm$ 3.4} \\
          & $F_1$      & \textbf{88.9 $\pm$ 2.7} & 84.8 $\pm$ 3.2 & 86.3 $\pm$ 3.1 & 88.3 $\pm$ 2.7 \\
\midrule
Pituitary & Precision & 91.9 $\pm$ 2.4 & 91.2 $\pm$ 2.5 & \textbf{93.7 $\pm$ 2.2} & 93.5 $\pm$ 2.1 \\
          & Recall    & 91.9 $\pm$ 2.5 & 91.9 $\pm$ 2.4 & 88.2 $\pm$ 2.9 & \textbf{96.3 $\pm$ 1.7} \\
          & $F_1$      & 91.9 $\pm$ 1.8 & 91.5 $\pm$ 1.8 & 90.8 $\pm$ 1.9 & \textbf{94.9 $\pm$ 1.4} \\
\midrule
Macro avg & Precision & 87.7 $\pm$ 1.6 & 84.5 $\pm$ 1.7 & 86.3 $\pm$ 1.5 & \textbf{89.4 $\pm$ 1.5} \\
          & Recall    & 88.1 $\pm$ 1.5 & 85.5 $\pm$ 1.6 & 85.2 $\pm$ 1.7 & \textbf{90.0 $\pm$ 1.4} \\
          & $F_1$      & 87.9 $\pm$ 1.5 & 84.9 $\pm$ 1.7 & 85.6 $\pm$ 1.6 & \textbf{89.6 $\pm$ 1.4} \\
\midrule
Weighted avg & Precision & 87.8 $\pm$ 1.5 & 84.9 $\pm$ 1.6 & 85.8 $\pm$ 1.5 & \textbf{89.8 $\pm$ 1.4} \\
             & Recall    & 87.8 $\pm$ 1.5 & 84.9 $\pm$ 1.6 & 85.3 $\pm$ 1.6 & \textbf{89.8 $\pm$ 1.4} \\
             & $F_1$      & 87.7 $\pm$ 1.5 & 84.9 $\pm$ 1.6 & 85.4 $\pm$ 1.6 & \textbf{89.8 $\pm$ 1.4} \\
\bottomrule
\end{tabular}
\normalsize
\end{table}

\begin{figure*}
    \centering
    \includegraphics[width=1.0\textwidth]{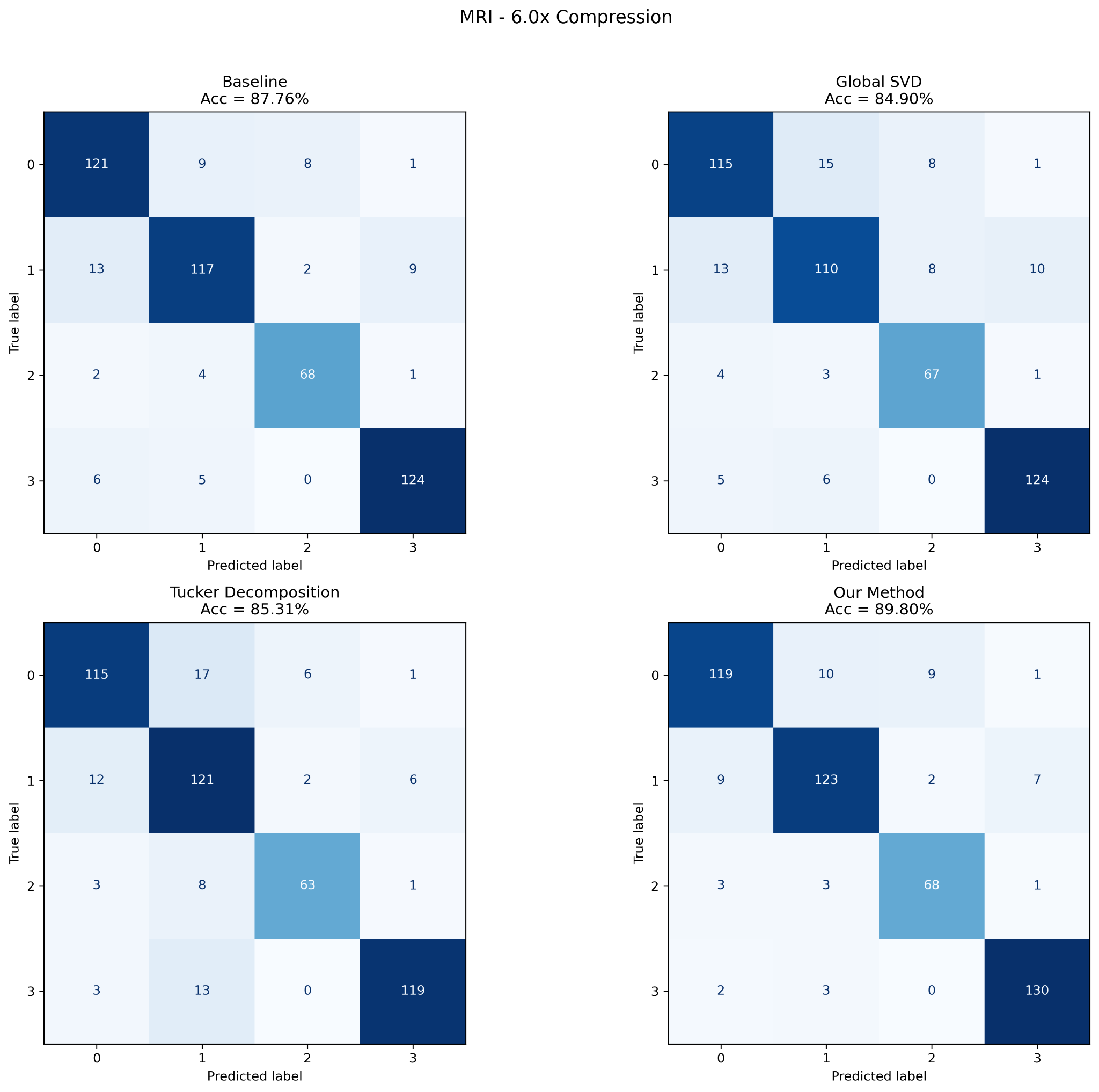}
    \caption{Confusion matrices for the baseline model, Global SVD, Tucker decomposition, and the proposed method at the \(6\times\) compression budget. The proposed method preserves the original class structure and achieves the highest classification accuracy under stronger compression.}
    \label{fig:confusion_6x}
\end{figure*}

The confusion matrices in Fig.~\ref{fig:confusion_6x} show that the proposed method preserves the overall structure of the original classifier even under stronger compression. Both Global SVD and Tucker decomposition exhibit substantially greater confusion between glioma and meningioma as the retained rank decreases. In contrast, the proposed method increases the number of correctly classified meningioma cases from \(117\) in the baseline to \(123\), and pituitary cases from \(124\) to \(130\), while leaving the no-tumour category almost unchanged. These findings suggest that the activation-guided clustering can be interpreted as a form of structured regularisation: by compressing different layers, spatial regions, and channel groups to different extents, the method removes redundant directions while preserving the local discriminative features necessary for accurate tumour classification.

\subsection{Hyper-parameter Trade-off Analysis}
To evaluate the behaviour of the proposed adaptive spatio-channel compression framework, we analyse its performance across four key hyper-parameters: the number of spatial clusters, $K_s$, the number of channel clusters, $K_c$, the energy threshold $\tau$ used for adaptive rank selection, and the maximum allowable rank $r_{\max}$. The number of spatial clusters controls how activation regions are partitioned, while the number of channel clusters determines the granularity of filter grouping within each spatial region. The energy threshold $\tau$ specifies the allowable proportion of spectral energy retained, and $r_{\max}$ sets an upper bound on the adaptive rank.

\begin{figure*}
    \centering
    \includegraphics[width=1.0\linewidth]{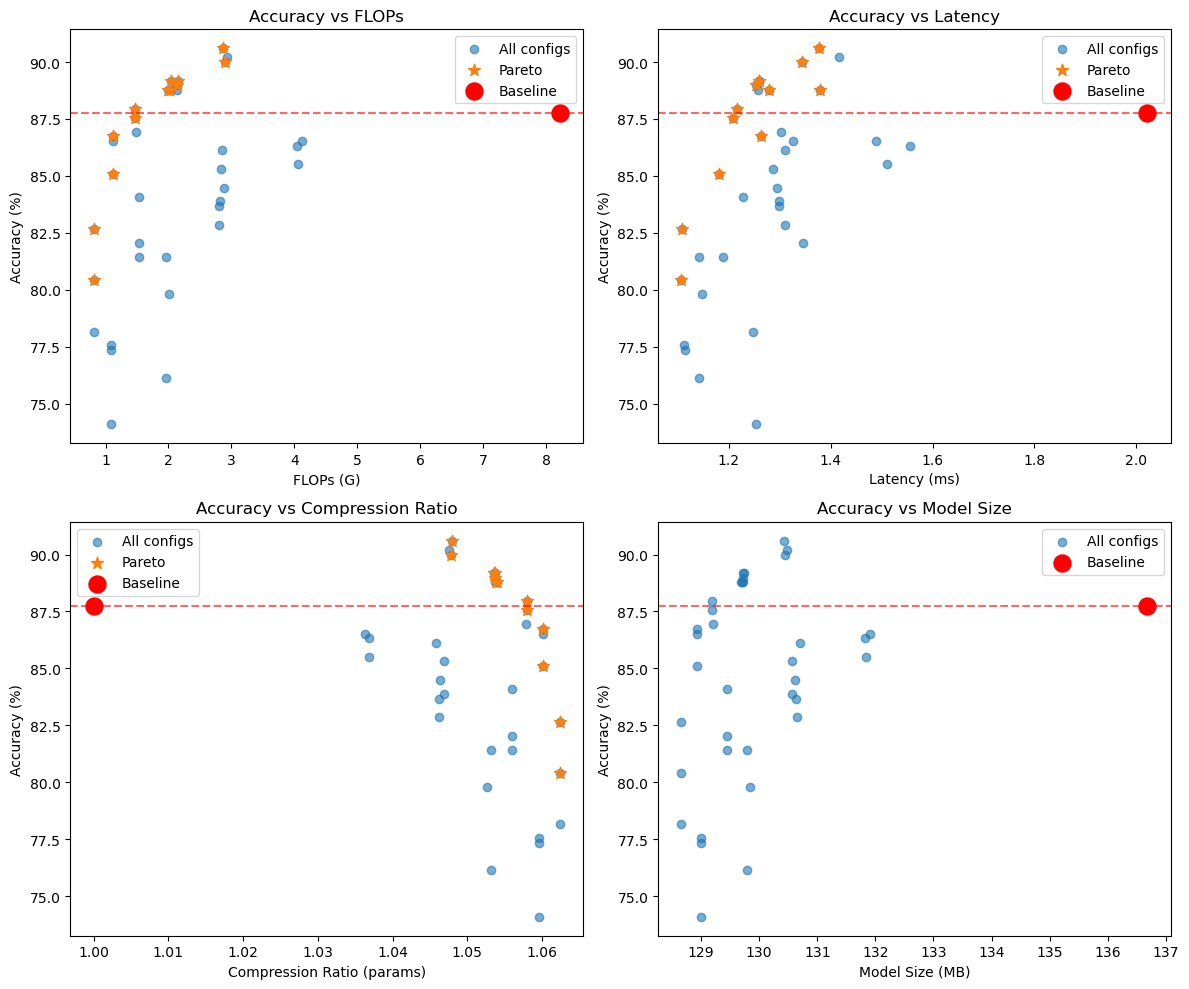}
    \caption{Pareto trade-off analysis across hyper-parameter configurations. Each point corresponds to a unique combination of $K_s$, $K_c$, $\tau$, and $r_{\max}$. Pareto-optimal configurations are highlighted. The baseline model is shown as a red circle, with the dashed line indicating baseline accuracy. The four panels illustrate trade-offs between accuracy and (top-left) FLOPs, (top-right) latency, (bottom-left) parameter compression ratio, and (bottom-right) model size.}
    \label{fig:pareto_tradeoff}
\end{figure*}

Fig.~\ref{fig:pareto_tradeoff} presents the trade-off behaviour across all evaluated configurations. Each point corresponds to a unique combination of these hyper-parameters, while Pareto-optimal configurations are highlighted. In general, increasing $K_s$ allows for a finer partitioning of activation regions, enabling a more localised low-rank approximation. Moderate values tend to provide a favourable balance between flexibility and compression, whereas excessively large values may reduce cluster sizes and limit effective rank estimation. Similarly, increasing $K_c$ allows more granular grouping of filters, which can improve reconstruction fidelity but may reduce compression efficiency if clusters become too small. The energy threshold $\tau$ directly controls the degree of truncation in the singular value spectrum. Smaller values yield stronger compression by discarding more singular components, resulting in lower FLOPs and latency. However, overly aggressive truncation may remove informative structure. Conversely, larger values retain more energy, resulting in higher accuracy but reduced computational gains.

The maximum rank $r_{\max}$ acts as a stabilising constraint on adaptive rank growth. Smaller values lead to stronger regularisation and compression, while larger values increase representational capacity. The results indicate that moderate values of $K_s$, $K_c$, $\tau$, and $r_{\max}$ provide a favourable balance between efficiency and accuracy. Importantly, the method maintains competitive accuracy across a substantial region of the hyper-parameter space, with performance generally degrading under more aggressive compression settings. Multiple configurations outperform the baseline model in accuracy while simultaneously reducing computational complexity. The Pareto frontier demonstrates that the framework provides several optimal trade-off points, enabling flexible selection according to deployment constraints without substantial degradation in predictive performance for well-chosen hyper-parameter settings.

\section{Discussion and Limitations}{\label{sec:discussion}}
The experimental results demonstrate that the proposed adaptive spatio‑channel low‑rank compression method achieves substantial computational savings while maintaining or improving classification performance at \(3\times\) and \(6\times\) compression budgets.  At the \(6\times\) budget, the compressed model reduces FLOPs from \(8.21\)\,G to \(1.55\)\,G (about \(81.1\%\) reduction) and achieves a \(1.38\times\) latency speed‑up, while increasing accuracy from \(87.76\%\) to \(89.80\%\) and raising the macro \(F_1\)-score from \(0.879\) to \(0.896\). The substantial reduction in FLOPs, despite a modest decrease in total parameter count (from \(35.83\)\,M to \(33.96\)\,M), reflects the structural characteristics of CNNs: convolutional layers dominate computational cost, whereas fully connected layers account for a large proportion of parameters. By exploiting redundancy in convolutional filters through activation‑guided clustering and adaptive rank truncation, the proposed method directly reduces the primary source of inference complexity.

The observed improvement in classification accuracy suggests that structured low‑rank decomposition, followed by fine‑tuning, appears to provide an additional regularising effect beyond standard weight decay. By removing redundant and highly correlated filter components, the compression reduces over‑parameterisation and constrains effective model capacity, which appears to enhance generalisation. The meningioma class benefits most notably, with its \(F_1\)-score increasing from \(0.848\) (baseline) to \(0.879\) at the \(6\times\) budget. Similarly, pituitary tumour classification improves from \(0.922\) to \(0.949\). The reduction in inter‑class confusion between glioma and meningioma, evident in Figs.~\ref{fig:confusion_3x} and~\ref{fig:confusion_6x}, further indicates that eliminating redundant structure increases discriminative representations.

Per‑layer analysis reveals that the proposed method adapts compression non‑uniformly according to layer depth. At the moderate \(3\times\) budget, \texttt{Conv2} is compressed most aggressively, by \(74.9\%\), indicating substantial redundancy in early filters. Under the tighter \(6\times\) budget, however, the method preserves more parameters in \texttt{Conv2}, with a reduction of \(75.6\%\), while applying stronger compression to \texttt{Conv4} at \(84.5\%\). This allocation suggests that low‑level features in earlier layers are more sensitive to over‑compression, whereas deeper layers contain increasingly correlated representations that tolerate aggressive rank reduction. In contrast, Global SVD and Tucker decomposition distribute compression nearly uniformly across layers at both budgets, lacking the ability to account for layer‑specific redundancy.

The hyper‑parameter trade‑off analysis demonstrates that the method is robust across a range of settings (as shown in Fig.~\ref{fig:pareto_tradeoff}). Multiple Pareto‑optimal configurations simultaneously improve accuracy and reduce computational cost, indicating that precise hyper‑parameter tuning is not critical for obtaining favourable trade‑offs. Moderate values of \(K_s\), \(K_c\), \(\tau\), and \(r_{\max}\) consistently yield configurations that outperform the baseline while achieving substantial FLOPs and latency reductions. This flexibility allows practitioners to select operating points according to deployment constraints without substantial degradation in predictive performance.

While the framework achieves significant computational savings, several practical considerations remain. The proposed method focuses exclusively on convolutional layers, where spatial and channel structures enable activation‑guided clustering and localised rank approximation. Fully connected layers are not compressed in the current implementation; as a result, overall parameter reduction is constrained by the backbone network's architectural characteristics. The clustering stage introduces a one‑time preprocessing cost during model transformation. Although this overhead does not affect inference‑time performance, it may become more noticeable when compressing very large networks or when exploring extensive hyper‑parameter grids. Furthermore, activation statistics are estimated from a subset of the training data, and the quality of clustering may depend on the representativeness of the sampled activations. In practice, this dependence is mitigated by using held‑out validation data and aggregating statistics over multiple batches, but it remains an important consideration if activation distributions shift substantially across data subsets or deployment domains.

\section{Conclusion}{\label{sec:conclusion}}
This paper presented a hierarchical spatio‑channel low‑rank compression framework that adaptively identifies and removes redundancy in CNNs by jointly considering spatial and channel coherence. The method partitions feature maps into spatial regions, groups channels by co‑activation patterns within each region, and applies rank‑adaptive SVD to each resulting cluster. This localised, activation‑guided approach contrasts with conventional low‑rank techniques, such as Global SVD and Tucker decomposition, which apply compression under a Global budget without explicitly modelling spatial‑channel coherence, resulting in nearly uniform reductions across layers.

Experiments on a brain tumour MRI classification task demonstrate that the proposed method achieves substantial computational savings while maintaining or improving classification accuracy. The framework distributes compression non-uniformly according to layer depth: at the \(3\times\) budget, \texttt{Conv2} is compressed most aggressively, whereas at the \(6\times\) budget the method preserves early layers and shifts stronger compression toward deeper layers, revealing increased redundancy in higher-level representations. In contrast, the baselines exhibit nearly uniform compression across all layers regardless of budget. Class‑wise evaluation, with bootstrap standard errors quantifying the finite‑sample uncertainty, shows consistent improvements for challenging categories such as meningioma, where the proposed method substantially outperforms both Global SVD and Tucker decomposition. The method also reduces FLOPs and inference latency significantly, offering a flexible Pareto frontier across a range of hyper‑parameter settings.

These findings suggest that activation‑guided structured compression, combined with fine‑tuning, imposes an implicit regularisation beyond standard weight decay. By discarding correlated filter components, the method reduces over‑parameterisation and improves generalisation.

The current implementation compresses only convolutional layers; fully connected layers remain uncompressed, limiting overall parameter reduction in architectures where they dominate. The clustering stage introduces a one‑time preprocessing cost that, while negligible at inference, may increase with larger networks or extensive hyper‑parameter search. Activation statistics are estimated from data subsets, and performance may vary if the deployment distribution differs substantially. Future work will extend the framework to deeper architectures such as ResNet-18 \cite{he2016deep} and VGG-16 \cite{simonyan2014very}, investigate compression of fully connected layers, assess performance across different datasets, and explore combinations with pruning and quantisation to further improve efficiency. 

%\section*{Acknowledgment}

%The preferred spelling of the word ``acknowledgment'' in American English is 
%without an ``e'' after the ``g.'' Use the singular heading even if you have 
%many acknowledgments. Avoid expressions such as ``One of us (S.B.A.) would 
%like to thank $\ldots$ .'' Instead, write ``F. A. Author thanks $\ldots$ .'' In most 
%cases, sponsor and financial support acknowledgments are placed in the 
%unnumbered footnote on the first page, not here.
\section{REFERENCES}
\bibliographystyle{IEEEtran}
\bibliography{sn-bibliography}
\end{document}